\documentclass{article}

\usepackage{arxiv}

\usepackage[utf8]{inputenc} % allow utf-8 input
\usepackage[T1]{fontenc}    % use 8-bit T1 fonts
\usepackage{hyperref}       % hyperlinks
\usepackage{url}            % simple URL typesetting
\usepackage{booktabs}       % professional-quality tables
\usepackage{amsfonts}       % blackboard math symbols
\usepackage{nicefrac}       % compact symbols for 1/2, etc.
\usepackage{microtype}      % microtypography
\usepackage{lipsum}		% Can be removed after putting your text content
\usepackage{graphicx}
\usepackage{natbib}
\usepackage{doi}
\usepackage{hyperref}
\usepackage{url}
\usepackage{graphicx}
\usepackage{tikz}
\usetikzlibrary{positioning}
\usepackage[linesnumbered]{algorithm2e}
\newcommand{\tp}{{t^{'} < t}}
\usepackage{wrapfig}
\usepackage{lipsum}
\usepackage{float}
\usepackage{caption}
\usepackage{subcaption}
\usepackage{amsmath}
\usepackage{amssymb}
\usepackage{booktabs}
\usepackage{array}
\newcolumntype{C}[1]{>{\centering\arraybackslash}p{#1}}

\usepackage{multirow}
\usepackage{makecell}
\usepackage{xcolor}
\newcommand*\samethanks[1][\value{footnote}]{\footnotemark[#1]}

\title{Mixture of Basis for Interpretable Continual Learning with Distribution Shifts}

\author{ Mengda Xu \thanks{Contributed equally} \\
	JPMorgan AI Research\\
    \texttt{\ mengda.xu@jpmorgan.com} \\
	\And
	Sumitra Ganesh\samethanks\\
	JPMorgan AI Research\\
    \texttt{\ sumitra.ganesh@jpmorgan.com} \\
	\And
	Pranay Pasula\\
	JPMorgan AI Research\\
	\texttt{\ pranay.pasula@jpmorgan.com} \\
}

\date{}

\renewcommand{\shorttitle}{\textit{arXiv} Template}

\begin{document}
\maketitle

\begin{abstract}
Continual learning in environments with shifting data distributions is a challenging problem with several real-world applications. In this paper we consider settings in which the data distribution (task) shifts abruptly and the timing of these shifts are not known. Furthermore, we consider a \textit{semi-supervised task-agnostic} setting in which the learning algorithm has access to both task-segmented and unsegmented data for offline training. We propose a novel approach called \textit{Mixture of Basis} models (MoB) for addressing this problem setting. The core idea is to learn a small set of \textit{basis models} and to construct a dynamic, task-dependent mixture of the models to predict for the current task. We also propose a new methodology to detect observations that are out-of-distribution with respect to the existing basis models and to instantiate new models as needed. We test our approach in multiple domains and show that it attains better prediction error than existing methods in most cases, while using fewer models than other multiple model approaches. Moreover, we analyze the latent task representations learned by MoB and show that similar tasks tend to cluster in the latent space and that the latent representation shifts at the task boundaries when tasks are dissimilar.
 
\end{abstract}
\section{Introduction}
Continual learning in environments with shifting data distributions is a challenging problem with several real-world applications e.g. weather and financial market data are known to have \emph{regime shifts}. In this paper, we consider settings where the data distribution (or \emph{task}) shifts abruptly but the timing of these shifts are not known. In such domains, segmenting historical data into tasks is often a difficult problem in itself and much of the data available is likely to be unsegmented. Motivated by this practical challenge, we consider a \textit{semi-supervised task-agnostic} setting in which the learning algorithm has access to both task-segmented and unsegmented data for offline training. Typically segmented data will be available only for a small set of tasks. In the online setting the algorithm does not observe task boundaries and encounters new tasks that were not present in the offline dataset.

For a learning algorithm to perform robustly in the presence of these distribution shifts, it must be able to recall previously encountered tasks as well as quickly adapt to new tasks by appropriately reusing elements from what was previously learned. 

\textbf{Contributions:} Our key contribution in this paper is a new approach to the aforementioned problem that is based on a mixture of \textit{basis models} (inspired by basis functions/vectors). The core idea is to learn a small set of basis models and to construct a dynamic, task-depdendent mixture of these models to predict for the current task. We also propose a new methodology to detect observations that are out-of-distribution with respect to the existing basis models and to instantiate new models. Our methodology uses a combination of model error and uncertainty to account for both covariate and concept shifts. Thus our approach allows for dynamic reuse of previously learned basis models across multiple tasks while simultaneously expanding the basis set \emph{as needed} to adapt to new tasks.

In our experiments on a synthetic regression domain and MuJoCo environments (HalfCheetah), we show that our approach, \textbf{M}ixture \textbf{o}f \textbf{B}asis \textbf{(MoB)}, achieves better mean-squared error (MSE) than comparable methods (MOLe \cite{nagabandi2019deep}, continuous adaptation, meta-learning based $k$-shot adaptation) in most cases. Compared to other \textit{multiple model} methods such as MOLe, our approach uses significantly fewer basis models indicating that our mixture-based approach allows for better reuse of previously learned models. 

Moreover, MoB learns interpretable task representations that can be especially useful in a task-agnostic setting to gain insight and confidence in the model's inference and prediction. We analyze the latent task representations learned by MoB and find that similar tasks tend to cluster together in the latent space and that the latent representation shifts at task boundaries when tasks are dissimilar. 

\section{Problem Statement}
\begin{figure}[hb]
\begin{center}
\scalebox{0.7}{
\begin{tikzpicture}[
lt/.style={circle, draw=black!60, fill=white!5, very thick, minimum size=12mm},
ob/.style={circle, draw=black!60, fill=gray!60, very thick, minimum size=12mm},
]
%Nodes
\node[lt]        (z0)       {$\mathcal{T}_{t-1}$};
\node[lt]        (z1)       [right=of z0]{$\mathcal{T}_t$};
\node[lt]        (z2)       [right=of z1]{$\mathcal{T}_{t+1}$};
\node[ob]        (x0)       [below=of z0] {$X_{t-1}$};
\node[ob]        (x1)       [below=of z1] {$X_{t}$};
\node[ob]        (x2)       [below=of z2] {$X_{t+1}$};
\node[ob]        (y0)       [below=of x0] {$Y_{t-1}$};
\node[ob]        (y1)       [below=of x1] {$Y_{t}$};
\node[ob]        (y2)       [below=of x2] {$Y_{t+1}$};

%Lines
\draw[->] (z0.east) -- (z1.west);
\draw[->] (z1.east) -- (z2.west);
\draw[->] (x0.south) -- (y0.north);
\draw[->] (x1.south) -- (y1.north);
\draw[->] (x2.south) -- (y2.north);
\path [black,bend left,->]   (z0) edge (y0);
\path [black,bend left,->]   (z1) edge (y1);
\path [black,bend left,->]   (z2) edge (y2);
% \draw[->] (maintopic.east) -- (rightsquare.west);
% \draw[->] (rightsquare.south) .. controls +(down:7mm) and +(right:7mm) .. (lowercircle.east);
\end{tikzpicture}
}
\caption{Generative model for non-stationary environment. The shaded variables (inputs $X_t$ and targets $Y_t$) are observed; the task $\mathcal{T}$ is unobserved.}
\label{Fig:GenModel}
\end{center}

\end{figure}
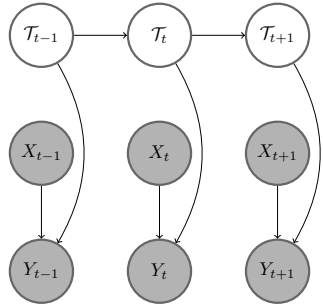

\begin{figure}
    \centering
    \hspace{-0.75cm}
    \includegraphics[width=\columnwidth]{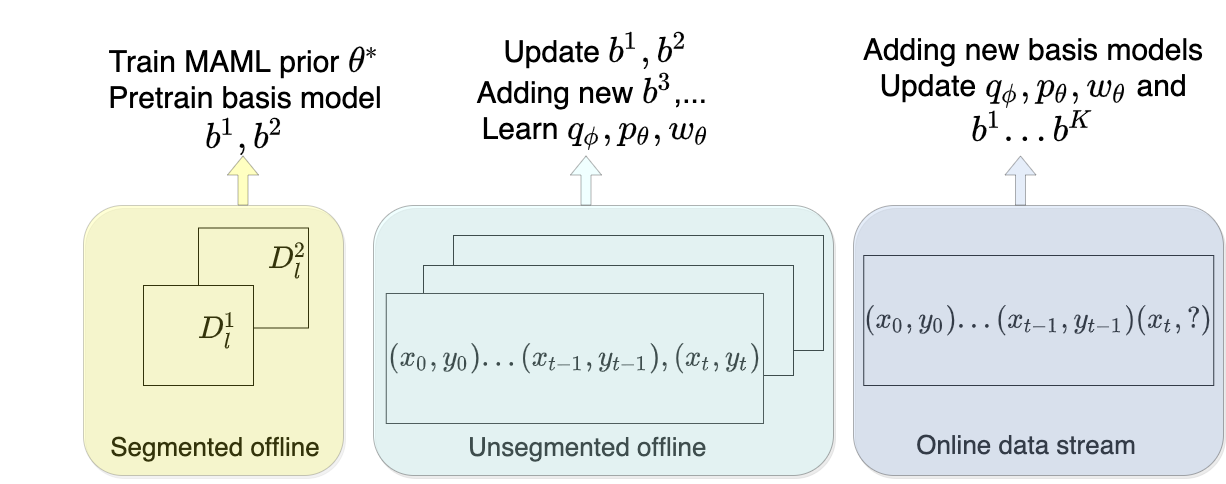}
    \caption{Offline and Online: overview of training and updates. The segmented dataset contains samples for $S$ tasks $\mathcal{D}_l := \{\mathcal{D}_l^1, \cdots, \mathcal{D}_l^S\}$ (here $S=2$). The unsegmented dataset contains trajectories $\tau = (x_{0:T}, y_{0:T})$.}
    \label{Fig:Datasets}
\end{figure}

Our goal is to learn a model that accurately predicts target variables $Y_t$ from inputs $X_t$ in a non-stationary environment in which the data distribution $P_{\mathcal{T}_t}(Y_t|X_t)$ can shift with time (depending on the task) and the task process $\mathcal{T}_t$ is Markov (see Fig.\ref{Fig:GenModel}).

We consider a \textit{semi-supervised task-agnostic} setting in which the learning algorithm has access to both task-segmented $\mathcal{D}_l$ and unsegmented data $\mathcal{D}_u$ for offline training (see Fig.\ref{Fig:Datasets}). Note that this is similar to most \textit{task-agnostic} continual learning approaches \citep{Caccia2020-ic, He2019-tz, He2021-yt, Jerfel2018-ou, nagabandi2019deep} that assume that task segmented data is available for offline training; the main difference is that we also leverage \textit{unsegmented} data.

The problems are: (a) \textbf{Offline training:} How do we effectively train the model using the offline dataset?, and (b) \textbf{Online adaptation:} How do we update the model in the online setting to adapt to task shifts and new tasks that were not in the offline dataset?

\section{Methods}

The core idea in our approach - Mixture of Basis models (MoB) - is that we could dynamically combine the predictions from a set of what we call \textit{basis models} to predict the target for the current task. 

\textbf{Basis Model Definition:} A \textit{basis model} is a model $b(X)$ that predicts $Y$ for a given $X$. With a slight abuse of notation, we will also use $b(Y|X)$ to denote the probability of $Y$ given $X$ under the basis model; point estimates for $Y$ are constructed by taking the expected value of this distribution. Given a set of $K$ basis models $\{ b^{(i)}\}_{i=1:K}$ and a latent task representation $Z_t \in \mathbb{R}^d$ for the current task $\mathcal{T}_t$, we express the conditional distribution of $Y_t$ given $X_t$ and $Z_t$ as the mixture
\begin{equation}
P(Y_t|X_t, Z_t) = \sum_{i=1}^{K} w^{i}(Z_t)~b^{(i)}(Y_t|X_t)
 \label{Eq:MoBform}
\end{equation}
where $\mathbf{w}(Z)$ maps the latent task representation to the standard simplex (i.e., $\sum_{i=1}^{K}  w^{i}(Z_t) = 1$ and $w^{i}(Z_t) \geq 0$ for $i=1,\dotsc, K$). Note that the mixing network $\textbf{w}()$ takes only the task representation as input, whereas the basis models are task-agnostic. We will first describe MoB's learning and inference procedure for a fixed number of basis models and then tackle the problem of when and how to add new basis models.

\begin{figure}[h!]
    \centering
    \includegraphics[scale=0.3]{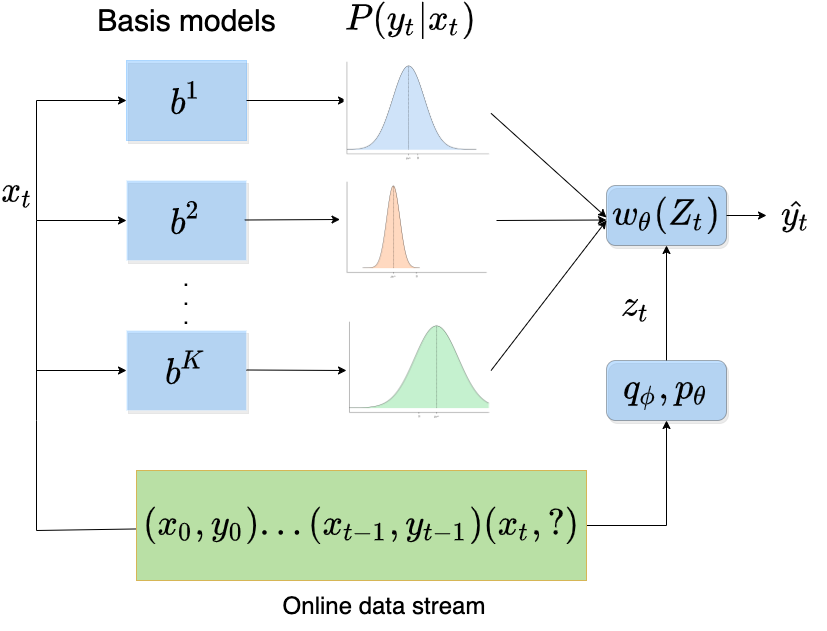}
    \caption{MoB pipeline.}
    \label{fig:mixture of basis}
\end{figure}

\subsection{Learning and Inference with a Fixed Basis Set}
Let's first consider that we have a fixed number of basis models, and would like to infer the latent task representations and make predictions, given a stream of observations $\tau := (x_{0:T}, y_{0:T})$ \footnote{\textbf{Notation}: We will use capital letters (e.g. $Z_t$) to denote random variables and small letters (e.g. $z_t$) to denote realizations/observations of the random variable}. 

 We use a sequential VAE construction (\cite{Chung2015-mr, Krishnan2015-db}) and learn the parameters $\theta$ and $\phi$ of the model $P$ and the inference model $q$, respectively, by maximizing the ELBO  $\mathcal{L}(\tau; \theta, \phi)$  w.r.t. $\theta$ and $\phi$ (see supplementary for full derivation).
\begin{align}
\label{Eq:ELBOfinal}
\begin{split}
 \mathcal{L}(\tau; \theta, \phi)  & = 
    E_{Z_t \sim q_{\phi}} \left[ \sum_{t=0}^{T} \log P_{\theta}(y_t| x_t, Z_t)\right] \nonumber \\
     & ~~- \sum_{t=1}^{T} E_{Z_{t-1} \sim q_{\phi}} [ D_{KL}(q_{\phi}( Z_{t} | Z_{t-1}, x_{t}, y_{t})\nonumber  \\
     & ~~~~~~~~~~~~~~~~~~~~~~~~~~~~~~~~~|| P_{\theta}(Z_t|Z_{t-1}) ] 
    % \\
    % &~~ - D_{KL}(q(Z_0) || P(Z_0))
\end{split}
\end{align}
The first term measures the  average reconstruction error using the model $P_{\theta}$ and the latent variables sampled from $q_{\phi}$, while the second term acts as a regularization constraining KL-divergence between the approximating posterior distribution $q_{\phi}$ and the prior over the task process $Z_t$. 
 
In order to compute the gradients w.r.t. the inference model parameters we use the re-parameterization trick in \cite{Kingma2013-hv} and estimate the ELBO as: 
% \begin{equation}
\begin{align}
%  \begin{split}
\hat{\mathcal{L}}(\tau; \theta, \phi) & =   \sum_{t=0}^{T} \log P_{\theta}(y_t| x_t, z_t)
 + \sum_{t=1}^{T}   \log P_{\theta}(z_t|z_{t-1}) \\
&  - \sum_{t=1}^{T}  \log q_{\phi}( z_{t} | z_{t-1}, x_{t}, y_{t}) ) \\
%  \end{split}
 \label{Eq:ELBOEst}
 \end{align}
% \end{equation}
where $z_t = \mu_{\phi}(z_{t-1}, x_t, y_t) + \sigma_{\phi}(z_{t-1}, x_t, y_t) \odot \epsilon $ and $\epsilon \sim \mathcal{N}(0, \mathbf{I})$. Note that the model $P_{\theta}$ here uses the mixture of basis formulation in Eq. \ref{Eq:MoBform} and has three components: (i) the mixture network $\mathbf{w}_{\theta}(Z)$, (ii) the basis models $b_{\theta}^{(i)}(Y|X)$, and (iii) the prior task model $p_{\theta}(Z_t|Z_{t-1})$.

\textbf{Pre-trained basis models:} In theory the basis models could be learned using the above procedure. However, we found in practice that this often led to the basis models being too similar to one another. To overcome this issue, we instead use the segmented task dataset $\mathcal{D}_l$ to pre-train a basis model for each task and then allow for adaptation.

\textbf{Uncertainty estimation using ensembles:} We would like to estimate the uncertainty in each basis model in order to assess when to instantiate new models. We use deep ensembles \citep{lakshminarayanan2017simple} for uncertainty estimation by training an ensemble of $M$ networks $\{b^{(i)}_j\}_{j=1:M}$ for each basis model $b^{(i)}$. Each model $b^{(i)}_j$ in the ensemble outputs the mean and variance of an isotropic Gaussian i.e.  $b^{(i)}_j (Y|X=x) = \mathcal{N}(\mu_{ij}(x),\sigma_{ij}(x))$. As in \cite{lakshminarayanan2017simple}, we weigh each model in the ensemble equally to construct the predictive distribution for the $i$-the basis as mixture of Gaussians,
\begin{equation}
b^{(i)}(Y_t|X_t) = \frac{1}{M} \sum_{j=1}^{M} b^{(i)}_j(Y_t|X_t)
\label{Eq:BasisEnsemble}
\end{equation}
and a point estimate 
\begin{equation}
\hat{y}_t = \frac{1}{M} \sum_{i=1}^{K} w^i(z_t) \sum_{j=1}^{M} \mu_{ij}(x_t)
\label{Eq:PointEst}
\end{equation}

\textbf{Fast basis instantiation using MAML:} During offline and online training, we might have only few data points available for instantiation of a basis model. To enable fast instantiation, we use a meta-learned prior using MAML \citep{DBLP:journals/corr/FinnAL17}. Since we need ensemble basis models, we train an ensemble MAML prior $\{\theta^*_j\}_{j=1}^{M}$ using the segmented task dataset and $b^{(i)}_j$ is adapted to task $\mathcal{T}_i$ from $\theta^*_j$ using the segmented data $\mathcal{D}_l^i$ for that task. 

\subsection{Adding New Basis Models}

For our approach to be able to adapt to new tasks in the offline or online setting, it needs to detect when the observations are out-of-distribution (OoD) with respect to the current set of models. Prediction errors \citep{He2021-yt} and model uncertainty \citep{Farquhar2018-ry} have been used to detect OoD samples or distribution shifts. In case of \emph{covariate} shift, where inputs $X$ are OoD with respect to the offline dataset, well-calibrated models would have high uncertainty. But in case of a \emph{concept} shift, the model could be confident in its predictions (it has seen similar inputs $X$ before) while still making errors because $P(Y|X)$ had changed. To reliably handle both types of shifts, we propose a novel \textbf{O}ut-of-\textbf{D}istribution \textbf{D}etection \textbf{S}core (ODDS) that combines model uncertainty and error. 

\begin{figure}[h!]
    \hspace*{-1cm}
    \centering
    \includegraphics[width=0.40\textwidth]{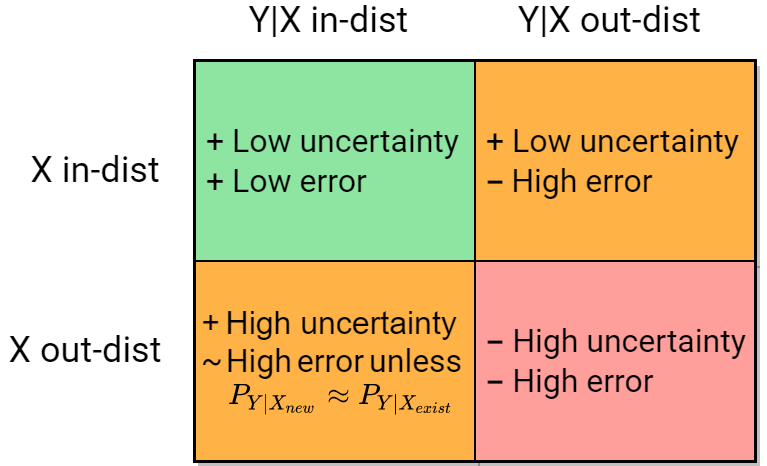}
    \caption{Model uncertainty and error for different scenarios of $X$ and $Y|X$ being in or out of distribution.}
    \label{fig:OODS}
\end{figure}

\RestyleAlgo{ruled}
\DontPrintSemicolon
\begin{algorithm}[t]
\textbf{Input}:$(x_t, y_t)$; meta-learned prior $\{\theta^*_j\}_{j=1}^{M}$\;
\If{$S_{ODDS}(x_t, y_t)>1$}{
Add $(x_t,y_t)$ into OOD buffer Q\; 
}
\If{$|Q|>L$}{

Adapt $b^{new}_j$ from $\theta^*_j$ to Q, $j=1,..,M$ \;
Add $b^{new}$ into mixture model.\; 
Clear buffer Q\; 
}
\caption{ODDS - Instantiating new basis}
\label{Algo:ODDS}
\end{algorithm}[t]

To decide whether a data point $(x, y)$ is OoD or not, we define a binary random variable $D$ that can take values $\{ I, O\}$ where $I$ and $O$ denote in and out of distribution, respectively. The decision on whether $(x, y)$ is OoD can then be based on the score 
\begin{equation}
    S_{ODDS}(x,y) := \frac{P(y|D=O,x)P(D=O|x)}{P(y|D=I,x)P(D=I|x)}
\end{equation}
with a default threshold of 1. Note that, just as with model uncertainty, $P(D|x)$ only depends on the input observation $x$. On the other hand, the likelihood $P(y|D,x)$ is evaluated after the ground truth $y$ is observed (similar to the model error).  

\textbf{Computing the likelihood $\mathbf{P(y|D, x)}$:} We determine the in-distribution likelihood as $P(y|D=I,x) = \max_i~b^{(i)}(y|x)$, by considering the basis model with the highest likelihood for the sample. To approximate the out-of-distribution likelihood $P(y|D=O,x)$, we create a new basis $b^{new}$ by adapting from the MAML prior $\theta^*$ using recently observed data and use the likelihood $b^{new}(y|x)$ under this model; the justification for this approximation is that if the data is OoD with respect to the current basis set, we need to fit a new model to evaluate the likelihood. 

\textbf{Computing the prior $\mathbf{P(D|x)}$:} To approximate the prior $P(D|x)$, we create a normalized uncertainty score for each basis and convert it into a probability-like measure. To create a normalized uncertainty score for the $i$-th basis, we normalize the total variance of the ensemble by the variance of the components 

\begin{equation}
    score_i(x)=  \frac{M^{-1}\sum\limits_{j=1}^{M}(\sigma_{ij}^2(x)+\mu_{ij}^2(x)) - \mu_{i}^2(x)}{M^{-1}\sum\limits_{j=1}^{M}\sigma_{ij}^2(x)}
\end{equation}
where $\mu_i(x) := M^{-1} \sum_{j=1}^{M} \mu_{ij}$. Intuitively, this score normalizes the total uncertainty by the aleatoric or \textit{irreducible} uncertainty $\sigma_{ij}^2$; so as the \textit{epistemic} uncertainty reduces and $\mu_{ij} \rightarrow \mu_i,~\forall j$ (the models in the ensemble overlap), the $score_i \rightarrow 1$. We convert the normalized score into a probability by exponential scaling,  
$
P(D=I|x) = e^{(1-score(x))/\eta}
$
where $\eta$ is a temperature parameter that modulates the sensitivity to the score (less sensitive as $\eta$ increases). To be conservative, we use the minimum score from all basis models for the $S_{ODDS}$ calculation. 

MoB maintains a buffer of OoD samples and instantiates a new basis when the buffer size exceeds a threshold. The procedure to add a new basis is summarized in Algorithm \ref{Algo:ODDS} and is used by the offline and online algorithms summarized in Algorithm \ref{Algo:Offline} and \ref{Algo:Online}.

\RestyleAlgo{ruled}
\DontPrintSemicolon
\begin{algorithm}[t]
\textbf{Input}: segmented data $\mathcal{D}_{l} = \{D_l^1,\cdots, D_l^S\}$; unsegmented data $\mathcal{D}_{u}$\; 

\textbf{Return}: $\{b_{\theta}^{(i)}\}$, $q_{\phi}$, $p_{\theta}$, $\textbf{w}_{\theta}$, $\theta^*$\; 

Train meta-learned prior $\{\theta^*_j\}_{j=1:M}$ using $\mathcal{D}_l$ \;
\For{i = 1:S}{
    Adapt $b^{(i)}_j$ to $D_l^i$ from $\theta^*_j$\, for $j=1,\cdots,M$
}

\While{Not converged}{
  Sample a minibatch of $B$ trajectories from $D_{u}$\;
  \For{each $\tau = \{x_{0:T}, y_{0:T}\}$ in minibatch}{
    Init $z_0$ \;%\sim \mathrm{Unif}(\mathbb{R}^d)$ \;
    \For{t=1:T}{
    Sample $\epsilon_t \sim \mathcal{N}(0, \mathbf{I})$ \;
    $z_t = \mu_{\phi}(z_{t-1}, x_t, y_t) + \sigma_{\phi}(z_{t-1}, x_t, y_t) \odot \epsilon_t$ \; 
    Add new basis using ODDS$(x_t,y_t)$\;

    }
    Compute $\hat{\mathcal{L}}(\tau; \theta, \phi)$ using Eq. (\ref{Eq:ELBOEst})\;
}
  Compute minibatch loss $\hat{\mathcal{L}}_{\theta, \phi} =  - \sum_{i=1}^{B} \hat{\mathcal{L}}(\tau^{(i)}; \theta, \phi)~~$\;
  Do gradient update: $\theta \leftarrow \theta - \nabla_{\theta} \hat{\mathcal{L}}_{\theta, \phi}; ~~~ \phi \leftarrow \phi - \nabla_{\phi} \hat{\mathcal{L}}_{\theta, \phi}$
  }
\caption{Mixture of Basis (MoB) - Offline}
\label{Algo:Offline}
\end{algorithm}

% MOB - ONLINE 
\RestyleAlgo{ruled}
\DontPrintSemicolon
\begin{algorithm}[t]

\textbf{Input}: online data stream $\tau$ \;
%\textbf{Start}:\;
Init $z_0$ \; % \sim \mathrm{Unif}(\mathbb{R}^d)$ \;
\For{each time step t}{
 Sample $\epsilon_t, \tilde{\epsilon}_t \sim \mathcal{N}(0, \mathbf{I})$ \;
Compute  $z_t = \mu_{\theta}(z_{t-1}) + \sigma_{\theta}(z_{t-1})\odot \epsilon_t$ \;
Compute $\hat{y_t}$ using Eq.(\ref{Eq:PointEst})\;
Update  $z_t = \mu_{\phi}(z_t, x_t, y_t) + \sigma_{\phi}(z_t, x_t, y_t)\odot \tilde{\epsilon}_t$\;
Compute $\hat{\mathcal{L}}(\tau_{1:t}; \theta, \phi)$ by plugging into Eq.(\ref{Eq:ELBOEst})\;
Do gradient update: $\theta \leftarrow \theta - \nabla_{\theta} \hat{\mathcal{L}}_{\theta, \phi}; ~~~ \phi \leftarrow \phi - \nabla_{\phi} \hat{\mathcal{L}}_{\theta, \phi}$s \;
Add new basis using $ODDS(x_t,y_t)$\;
}
\caption{Mixture of Basis (MoB) - Online}
\label{Algo:Online}
\end{algorithm}

\section{Experiments}
Our goal is to study the following three aspects of our proposed approach: 
\begin{enumerate}
    \item \textbf{Performance}:  How does our approach compare in terms of average prediction error (MSE) over an online stream of tasks?
    \item \textbf{Scaling}: How many basis models does our approach (MoB) instantiate?
    \item \textbf{Interpretability}: Does the latent representation of the task ($Z$) reflect the similarity of tasks e.g. do the latent representations for similar tasks cluster together? Are the task shifts reflected in shifts in the latent space? 
\end{enumerate}

We evaluate our method on a synthetic regression domain and on building environment models for HalfCheetah (MuJoCo \cite{DBLP:journals/corr/abs-1905-10681}). \footnote{Note that though the MuJoCo environment (HalfCheetah) is typically used in an reinforcement learning setting, here we are only interested in a regression task constructed using this domain (i.e., the task of predicting the next state, given the current state and action.)} In both domains we define multiple tasks as described below. A subset of the tasks are included in the segmented and unsegmented offline dataset (task partitions are described in supplementary). After training on the offline dataset, all approaches are evaluated on an online data stream generated by a Markovian task process with a fixed, uniform probability per time step of switching to a different task. In particular, the online stream contains tasks that were not present in the offline data set in order to test the ability of the approaches to adapt to new tasks. For each domain and task partition, we report results on 10 different seeds.
\begin{figure}[h]
\begin{subfigure}{0.45\textwidth}
    \begin{center}
        \includegraphics[scale=0.3]{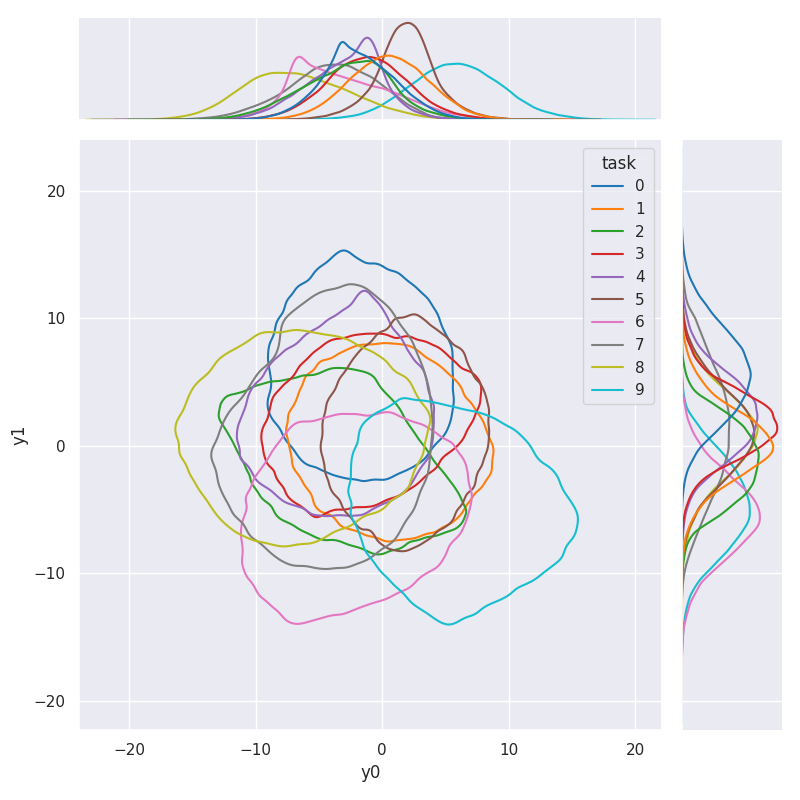}
    \end{center}
    \vspace{.05in}
    \caption{\textbf{Regression.} Marginal and joint probability density function estimates of variable $y$ for each task in the synthetic Regression domain. For clarity only one level set per task is shown.}
    \label{fig:reg_densities}
\end{subfigure}
\hfill
\begin{subfigure}{0.45\textwidth}
    \begin{center}
        \includegraphics[scale=0.30]{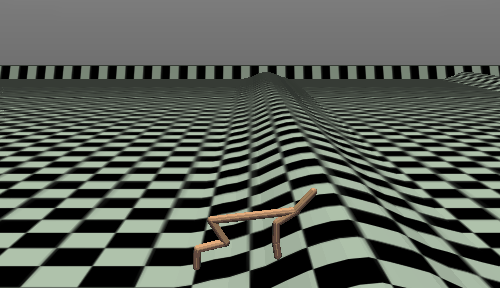}
    \end{center}
    \vspace{.05in}
    \caption{\textbf{HalfCheetah ice-slope}, shown climbing a slope}
    \label{fig:half-cheetah_domain_fig}
\end{subfigure}
\caption{Experiment domains.}
\end{figure}

\textbf{Regression:} We use randomly initialized neural networks to create $10$ different regression tasks. The input observation $X$ is sampled uniformly at random in $[-1, 1]$ and $Y \sim \mathcal{N}(\mu_i(x); \sigma_i(x))$ for the $i$-th task, where $\mu_i$ and $\sigma_i$ are randomly initialized networks. The advantage of creating regression tasks in this manner is that (a) the regression tasks are not overly simplistic, and (b) we can construct informative and well-understood dissimilarity measures (e.g. Kullback-Leibler divergence,  Bhattacharyya distance) between the conditional distributions $P(Y|X)$ under different tasks to quantify task similarity.

\textbf{HalfCheetah foot:} We create different tasks in the HalfCheetah environment by clipping the action space of one or two actuators to one-third of the original.  To generate the data under for different tasks, we rollout a policy trained in the standard HalfCheetah environment (using SAC \citep{DBLP:journals/corr/abs-1812-05905}) under the different task settings. Specifically, we clip the front foot(ff), front thigh(ft), back foot(bf), and back thigh(bt) to create 8 tasks - bt, ff, ftff, bfft, btff, btft, bfff and btbf. We randomly pick two tasks to be included in the segmented offline data $\mathcal{D}_l$ and four tasks (including the ones in $\mathcal{D}_l$) to be included in the unsegmented offline data set $\mathcal{D}_u$. All tasks are included in the online data stream. 

\textbf{HalfCheetah ice-slope:} We create another set of tasks using the HalfCheetah environment by changing the slopes and friction of the terrain. We create 4 different tasks for the offline training: 1) flat terrain with low friction / ice, 2) medium slope - medium slope with normal friction, 3) mixed (easy, medium, and hard) slopes with normal friction, and 4) medium slope with ice terrain. The data generation process is similar to the HalfCheetah foot. We randomly pick two tasks for the segmented offline dataset $\mathcal{D}_l$ and an additional two tasks for unsegmented offline dataset $\mathcal{D}_u$ (see supplementary for details). During the online phase, we always test our method in a low friction terrain with different slopes (mixed ice) which was never seen during offline training. \footnote{The construction of tasks here is limited by the technical difficulty of not being able to switch the friction easily for a Mujoco environment within a single rollout.}

We compare our method with three baselines - all of which use a meta-learned prior \citep{DBLP:journals/corr/FinnAL17} trained using offline data:
\begin{itemize}
    \item \textbf{MOLe:} MOLe \citep{nagabandi2019deep} is an online algorithm that instantiates multiple \textit{task} models from a meta-learned prior. We add an offline training phase to MOLe for fair comparison by running MOLe on each trajectory in the offline data  and carrying over any generated models. 
    \item \textbf{MAML k-shot:} This approach uses a single prediction model that is adapted from MAML prior $\theta^*$ using latest $k$ data points. 
    \item \textbf{MAML continuous:} This approach uses a single prediction model that is initialized using a meta-learned prior $\theta^*$ (learned using offline data). At each time step, the current model is updated using the most recent observations. 

\end{itemize}
While MAML k-shot and continuous optimize for adaptation to the most recent task, MOLe also tries to optimize recall of previous tasks by maintaining multiple task models. These baselines provide a useful contrast against our approach which also tries to achieve adaptation to new tasks by reusing previously learned basis models.
\begin{figure}[h!]
\begin{center}
%\framebox[4.0in]{$\;$}]
\vspace{-0.1cm}
\includegraphics[width=0.3\textwidth]{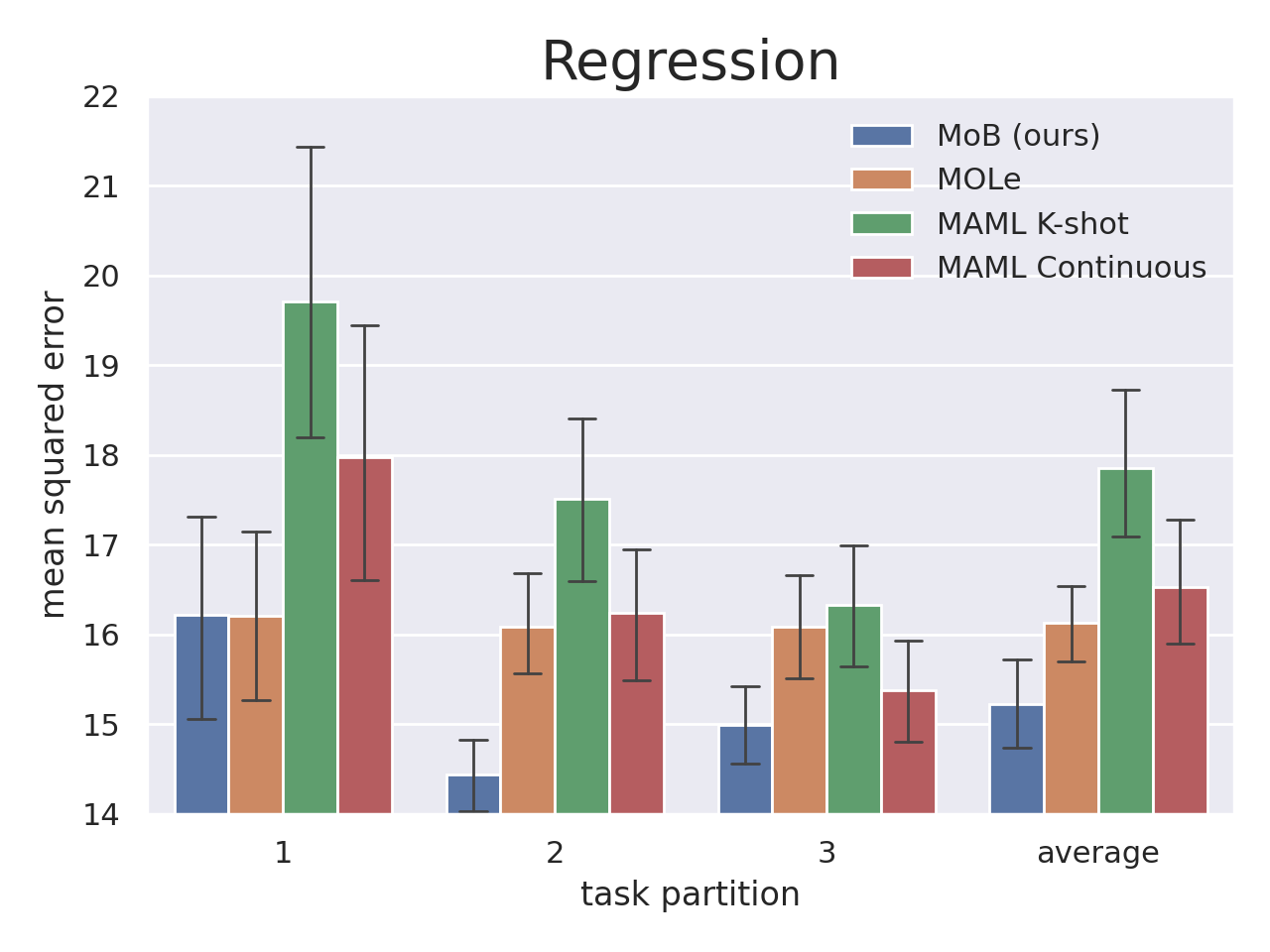}
\includegraphics[width=0.3\textwidth]{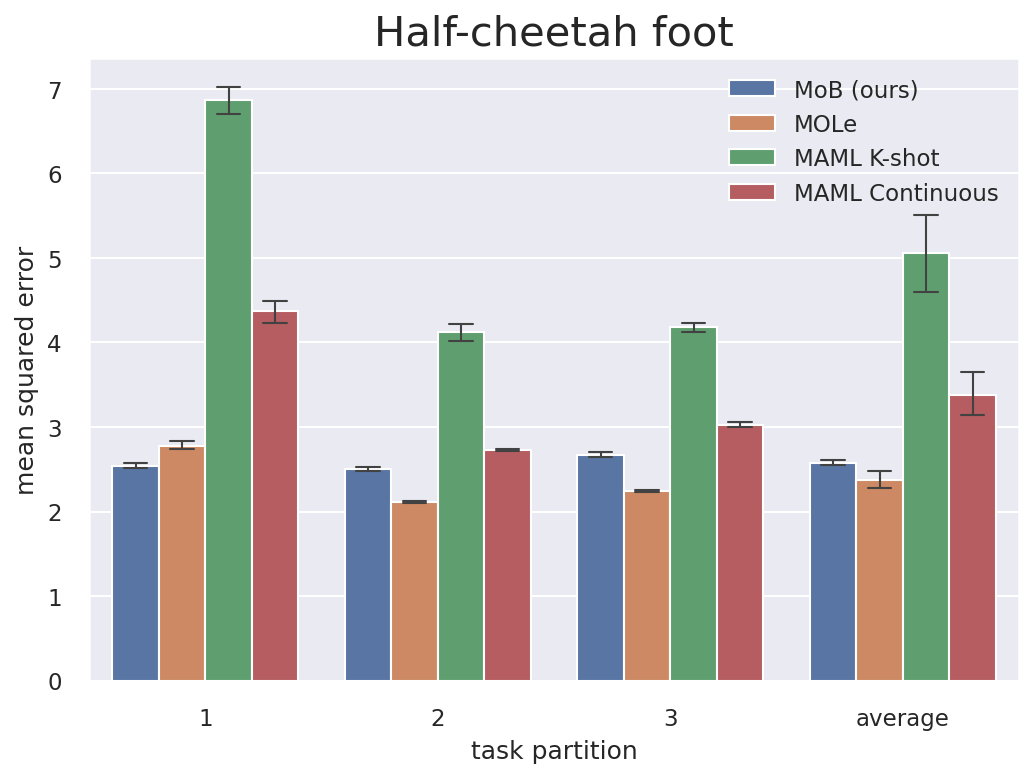}
\includegraphics[width=0.3\textwidth]{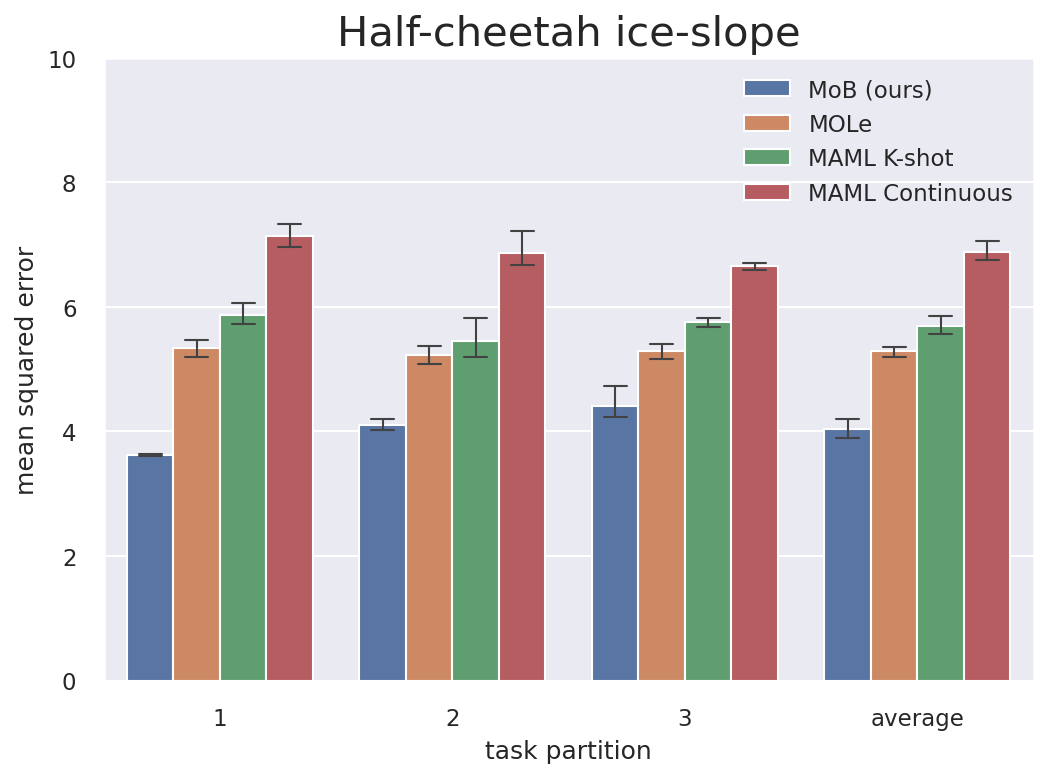}
\end{center}
\vspace{-0.2cm}
\caption{MSEs and 95\% confidence intervals over 10 seeds for each algorithm on each task partition.}
\end{figure}

\textbf{Performance and Scaling:} MoB achieves the lowest mean squared-error (MSE) on the online stream of tasks, in most task partitions and across domains. On both the regression task and HalfCheetah ice-slope, MoB achieves significantly better performance than all baselines while instantiating significantly fewer models than MOLe. For example, in the HalfCheetah ice-slope domain, MoB instantiates $13, 15$ and $2$ basis models (on average) in the three task partitions, in comparison to the $26, 53$ and $9$ task models instantiated by MOLe. In the regression domain, MoB instantiates $3$ models vs. MOLe's average of $12$ (details in Table \ref{tab:num_models}). This demonstrates MoB's ability to reuse previously learned basis models in order to achieve comparable or superior performance.  

In HalfCheetah foot, MoB only outperforms MOLe in the first task partition which contains segmented tasks clipping only one actuator; the other two have the segmented tasks clipping two actuators. To some degree, the first task partition provides more 'primitive' segmented tasks in the offline training phase. This opens up the interesting question and direction of whether we could design a curriculum of primitive tasks during offline training for MoB to enable better adaptation in the online phase. 

\begin{table}[b]
\caption{Average number of basis (task) models instantiated over 10 different seeds per partition.}
\begin{center}
\begin{tabular}{l|l|l|l}
\textbf{Domain} & \textbf{Partition} & \textbf{MoB} & \textbf{MOLe} \\
\hline
& 1 & \textbf{3.4} & 12.0\\ 
Regression & 2 & \textbf{3.6} & 11.8\\ 
& 3 & \textbf{3.3} & 11.6\\
\hline
& 1 & 3.0 & \textbf{2.3}\\ 
HalfCheetah foot & 2 & \textbf{3.0} & 3.9\\ 
& 3 & \textbf{3.0} & 3.4 \\
\hline

& 1 & \textbf{13.4} & 26.5\\ 
HalfCheetah ice-Slope & 2 & \textbf{14.8} & 52.8\\ 
& 3 & \textbf{2.0} & 9.2 \\
\bottomrule
\end{tabular}
\end{center}
 \label{tab:num_models}
\end{table}

\begin{figure}[h!]
\centering
\vspace{-0.1cm}

    \centering
        \includegraphics[width=0.45\textwidth]{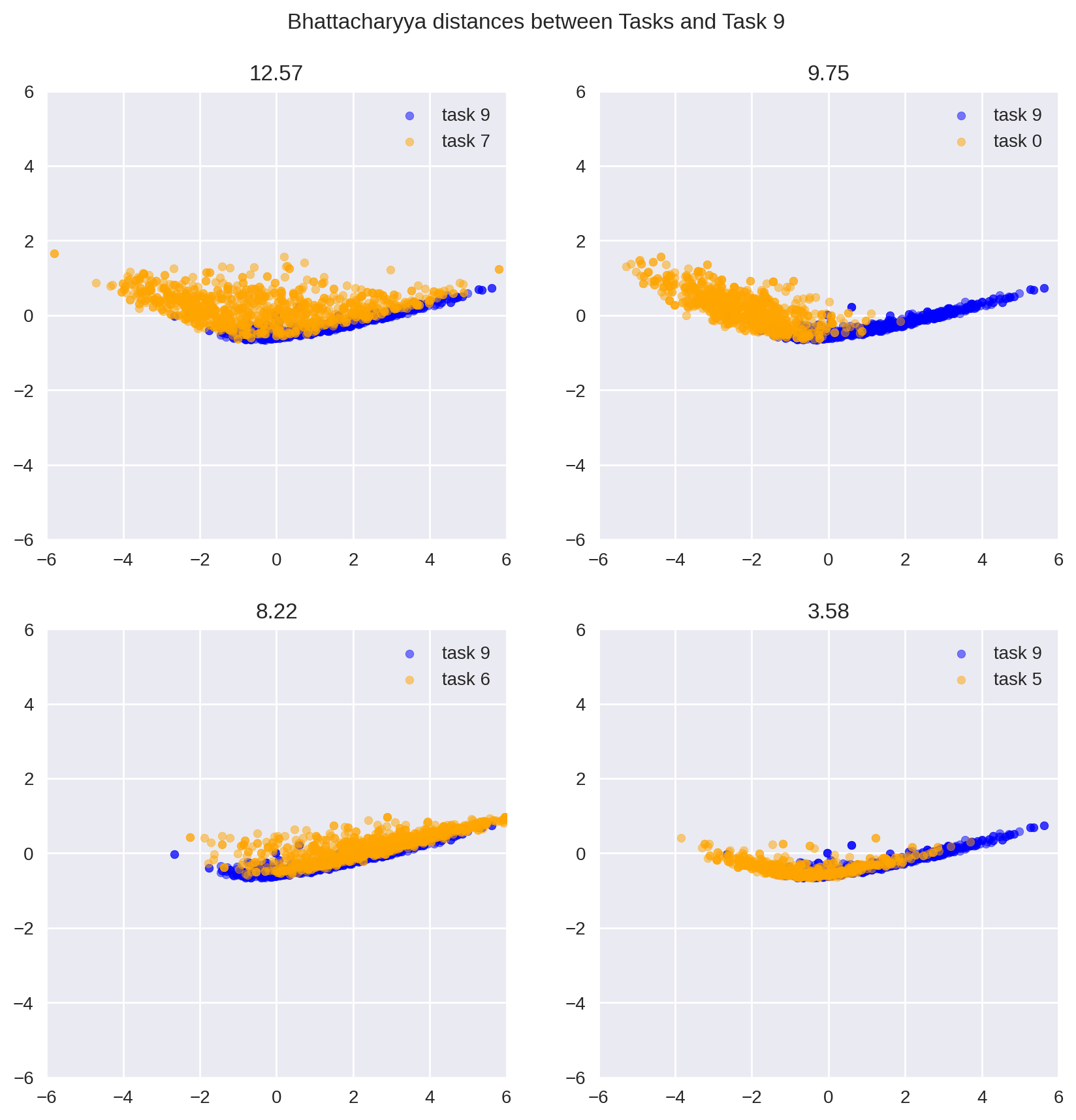}
    \caption{\textbf{Regression.} First two principal components of the $32$-dimensional latent space $Z$. Bhattacharyya distance between tasks is above each subplot; plots are ordered in terms of decreasing distance (increasing task similarity) to a reference task $9$ shown in blue.}
    \label{fig:pca_bhatta}

\end{figure}

\textbf{Interpretability:} We analyzed the principal components of the $Z_t$ inferred by MoB to study whether similar tasks cluster together in latent space. In the regression domain, we can quantify the task similarity by the Bhattacharya distance between the distribution of $Y$ under different tasks. As we can see in Fig. \ref{fig:pca_bhatta}, latent task representations are well separated if tasks are sufficiently different and overlap more as the similarity increases. We also confirmed that the trajectories of the $Z$-components with the highest variance typically show shifts that are aligned with the task boundaries if the tasks are dissimilar (see Fig. \ref{fig:z_space}; additional plots can be found in the supplementary material).

\begin{figure}[h!]
\centering
\vspace{-0.1cm}

    \centering
        \includegraphics[width=0.35\textwidth]{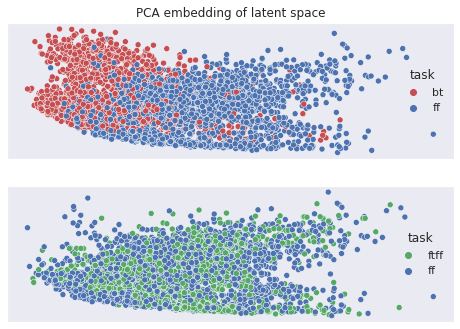}
    \caption{\textbf{Half-cheetah foot.} First two principal components of the $32$-dimensional latent space $Z$. The tasks \textit{ff} and \textit{bt} (top plot) are largely separated in the principal component space, while the tasks \textit{ff} and \textit{ftff} overlap with each other, likely  because the front foot \textit{ff} is clipped in both.} 
    \label{fig:pca_foot}

\end{figure}

\begin{figure}[h!]
    \begin{center}
        \includegraphics[width=0.45\textwidth]{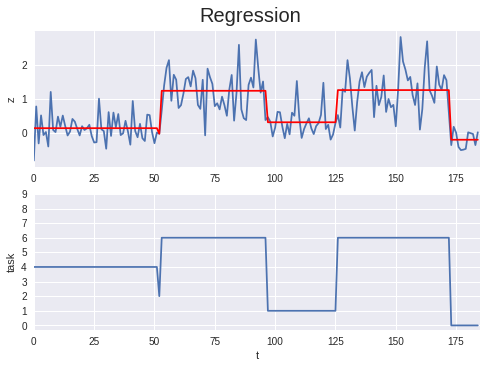}
    \end{center}
    \caption{\textbf{Regression.} Activations of a randomly selected $z$ component ($z_{15}$) over a randomly selected trajectory. Mean values of task segments are shown in red. The activations of all other 31 $z$ components is shown in Appendix.}
    \label{fig:reg_z15}
\end{figure}

\begin{figure}[h!]
    \centering
        \includegraphics[width=0.45\textwidth]{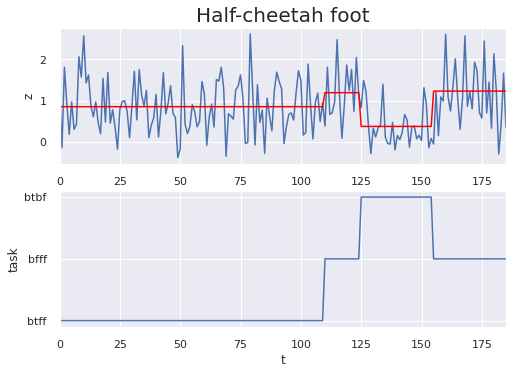}
    \caption{\textbf{HalfCheetah Foot.} $z$ component with maximum variance (mean values in red). When the task switches from \textit{btff} to \textit{bfff}, the activation doesn't change much, likely because \textit{ff} is present in both. However, it shifts significantly when the task changes to \textit{back}.}
    \label{fig:z_space}
\end{figure}

\section{Related Work}
 
Regularization and rehearsal based approaches to continual learning \citep{Maltoni2018-qo} typically use a single network and employ strategies to prevent catastrophic forgetting of previously learned tasks. Architectural approaches such as \cite{Rusu2016-kt} instantiate a new networks for each task but scale poorly with the number of tasks and are not applicable in task-agnostic settings. 

Recent works in continual learning  \citep{Finn2019-ke, Caccia2020-ic, Gupta2020-of}, leverage advances in meta-learning (e.g. MAML \citep{DBLP:journals/corr/FinnAL17}) to enable fast adaptation to the current task, instead of attempting to strictly remember previous tasks with no adaptation. However, these approaches don't allow for modularization and interpretability. A handful of works have proposed ways to combine model components that are different from our approach. \cite{He2019-tz} combines model components in parameter space whereas \cite{Jerfel2018-ou} proposes an approach where the MAML prior itself is modeled as a mixture distribution. MOLe \citep{nagabandi2019deep}, similar to our approach, creates multiple prediction models by leveraging MAML and is hence used as a baseline for comparison.

In MoB, we try to strike a balance between adaptability and interpretability by leveraging MAML for fast instantiation of basis models. Our approach of separating the task-specific and task-agnostic model components is similar to the "what and how" framework proposed in  \cite{He2019-tz}, and indeed can be thought of as a specific implementation.

Several methods to detect OoD samples have been proposed in the context of classification tasks - \cite{DBLP:journals/corr/HendrycksG16c} take the maximum of softmax probability as the confidence score to detect OoD samples while \cite{lee2018simple} define the score based on the Mahalanobis distance; \cite{DBLP:journals/corr/abs-1812-04606} and \cite{DBLP:journals/corr/abs-2104-03829} assume the models have access to a large data set of known outliers. \cite{He2019-tz} focus on a regression setting and use prediction errors with set thresholds to detect \textit{novel} samples. In contrast, we cast the detection problem as a binary decision and handle combinations of shifts in $P(X)$ and $P(Y|X)$ by jointly considering the model uncertainty and likelihood.

\section{Conclusions}

We propose a new approach, Mixture of Basis (MoB), that can learn robustly in the presence of distribution shifts. Our approach learns a set of basis models and constructs a dynamic, task-dependent mixture of the existing basis models to predict for the current task. We also introduce a new methodology (ODDS) to instantiate new basis models which takes into account both the uncertainty of the existing models before the ground truth is observed, as well as the likelihood of the ground truth after it is observed. We show in our experiments that MoB is able to perform better than comparable methods like MOLe in most cases, while instantiating significantly fewer models. Moreover, our analysis of the latent task representations learned by MoB indicate that it learns \textit{interpretable} task representations - we show that similar tasks cluster together in the latent space and that the latent representation shifts at task boundaries when the tasks are dissimilar. Thus, our proposed approach learns meaningful task representations while achieving good performance and scaling 

\paragraph{Disclaimer}
This paper was prepared for informational purposes by
the Artificial Intelligence Research group of JPMorgan Chase \& Co\. and its affiliates (``JP Morgan''),
and is not a product of the Research Department of JP Morgan.
JP Morgan makes no representation and warranty whatsoever and disclaims all liability,
for the completeness, accuracy or reliability of the information contained herein.
This document is not intended as investment research or investment advice, or a recommendation,
offer or solicitation for the purchase or sale of any security, financial instrument, financial product or service,
or to be used in any way for evaluating the merits of participating in any transaction,
and shall not constitute a solicitation under any jurisdiction or to any person,
if such solicitation under such jurisdiction or to such person would be unlawful.
\bibliographystyle{unsrtnat}
\bibliography{references}  %%% Uncomment this line and comment out the ``thebibliography'' section below to use the external .bib file (using bibtex) .

\begin{thebibliography}{21}
\providecommand{\natexlab}[1]{#1}
\providecommand{\url}[1]{\texttt{#1}}
\expandafter\ifx\csname urlstyle\endcsname\relax
  \providecommand{\doi}[1]{doi: #1}\else
  \providecommand{\doi}{doi: \begingroup \urlstyle{rm}\Url}\fi

\bibitem[Nagabandi et~al.(2019)Nagabandi, Finn, and Levine]{nagabandi2019deep}
Anusha Nagabandi, Chelsea Finn, and Sergey Levine.
\newblock Deep online learning via meta-learning: Continual adaptation for
  model-based rl, 2019.
\newblock URL \url{http://arxiv.org/abs/1812.07671}.

\bibitem[Caccia et~al.(2020)Caccia, Rodriguez, Ostapenko, Normandin, Lin,
  Caccia, Laradji, Rish, Lacoste, Vazquez, and Charlin]{Caccia2020-ic}
Massimo Caccia, Pau Rodriguez, Oleksiy Ostapenko, Fabrice Normandin, Min Lin,
  Lucas Caccia, Issam Laradji, Irina Rish, Alexandre Lacoste, David Vazquez,
  and Laurent Charlin.
\newblock Online fast adaptation and knowledge accumulation: a new approach to
  continual learning.
\newblock March 2020.
\newblock URL \url{http://arxiv.org/abs/2003.05856}.

\bibitem[He et~al.(2019)He, Sygnowski, Galashov, Rusu, Teh, and
  Pascanu]{He2019-tz}
Xu~He, Jakub Sygnowski, Alexandre Galashov, Andrei~A Rusu, Yee~Whye Teh, and
  Razvan Pascanu.
\newblock Task agnostic continual learning via meta learning.
\newblock June 2019.
\newblock URL \url{http://arxiv.org/abs/1906.05201}.

\bibitem[He and Sick(2021)]{He2021-yt}
Yujiang He and Bernhard Sick.
\newblock {CLeaR}: An adaptive continual learning framework for regression
  tasks.
\newblock January 2021.
\newblock URL \url{http://arxiv.org/abs/2101.00926}.

\bibitem[Jerfel et~al.(2018)Jerfel, Grant, Griffiths, and
  Heller]{Jerfel2018-ou}
Ghassen Jerfel, Erin Grant, Thomas~L Griffiths, and Katherine Heller.
\newblock Reconciling meta-learning and continual learning with online mixtures
  of tasks.
\newblock December 2018.
\newblock URL \url{http://arxiv.org/abs/1812.06080}.

\bibitem[Chung et~al.(2015)Chung, Kastner, Dinh, Goel, Courville, and
  Bengio]{Chung2015-mr}
Junyoung Chung, Kyle Kastner, Laurent Dinh, Kratarth Goel, Aaron~C Courville,
  and Yoshua Bengio.
\newblock A recurrent latent variable model for sequential data.
\newblock In C~Cortes, N~D Lawrence, D~D Lee, M~Sugiyama, and R~Garnett,
  editors, \emph{Advances in Neural Information Processing Systems 28}, pages
  2980--2988. Curran Associates, Inc., 2015.
\newblock URL
  \url{http://papers.nips.cc/paper/5653-a-recurrent-latent-variable-model-for-sequential-data.pdf}.

\bibitem[Krishnan et~al.(2015)Krishnan, Shalit, and Sontag]{Krishnan2015-db}
Rahul~G Krishnan, Uri Shalit, and David Sontag.
\newblock Deep kalman filters.
\newblock November 2015.
\newblock URL \url{http://arxiv.org/abs/1511.05121}.

\bibitem[Kingma and Welling(2013)]{Kingma2013-hv}
Diederik~P Kingma and Max Welling.
\newblock {Auto-Encoding} variational bayes.
\newblock December 2013.
\newblock URL \url{http://arxiv.org/abs/1312.6114v10}.

\bibitem[Lakshminarayanan et~al.(2017)Lakshminarayanan, Pritzel, and
  Blundell]{lakshminarayanan2017simple}
Balaji Lakshminarayanan, Alexander Pritzel, and Charles Blundell.
\newblock Simple and scalable predictive uncertainty estimation using deep
  ensembles, 2017.

\bibitem[Finn et~al.(2017)Finn, Abbeel, and
  Levine]{DBLP:journals/corr/FinnAL17}
Chelsea Finn, Pieter Abbeel, and Sergey Levine.
\newblock Model-agnostic meta-learning for fast adaptation of deep networks.
\newblock \emph{CoRR}, abs/1703.03400, 2017.
\newblock URL \url{http://arxiv.org/abs/1703.03400}.

\bibitem[Farquhar and Gal(2018)]{Farquhar2018-ry}
Sebastian Farquhar and Yarin Gal.
\newblock Towards robust evaluations of continual learning.
\newblock May 2018.
\newblock URL \url{http://arxiv.org/abs/1805.09733}.

\bibitem[Qureshi et~al.(2019)Qureshi, Johnson, Qin, Boots, and
  Yip]{DBLP:journals/corr/abs-1905-10681}
Ahmed~Hussain Qureshi, Jacob~J. Johnson, Yuzhe Qin, Byron Boots, and Michael~C.
  Yip.
\newblock Composing ensembles of policies with deep reinforcement learning.
\newblock \emph{CoRR}, abs/1905.10681, 2019.
\newblock URL \url{http://arxiv.org/abs/1905.10681}.

\bibitem[Haarnoja et~al.(2018)Haarnoja, Zhou, Hartikainen, Tucker, Ha, Tan,
  Kumar, Zhu, Gupta, Abbeel, and Levine]{DBLP:journals/corr/abs-1812-05905}
Tuomas Haarnoja, Aurick Zhou, Kristian Hartikainen, George Tucker, Sehoon Ha,
  Jie Tan, Vikash Kumar, Henry Zhu, Abhishek Gupta, Pieter Abbeel, and Sergey
  Levine.
\newblock Soft actor-critic algorithms and applications.
\newblock \emph{CoRR}, abs/1812.05905, 2018.
\newblock URL \url{http://arxiv.org/abs/1812.05905}.

\bibitem[Maltoni and Lomonaco(2018)]{Maltoni2018-qo}
Davide Maltoni and Vincenzo Lomonaco.
\newblock Continuous learning in {Single-Incremental-Task} scenarios.
\newblock June 2018.
\newblock URL \url{http://arxiv.org/abs/1806.08568}.

\bibitem[Rusu et~al.(2016)Rusu, Rabinowitz, Desjardins, Soyer, Kirkpatrick,
  Kavukcuoglu, Pascanu, and Hadsell]{Rusu2016-kt}
Andrei~A Rusu, Neil~C Rabinowitz, Guillaume Desjardins, Hubert Soyer, James
  Kirkpatrick, Koray Kavukcuoglu, Razvan Pascanu, and Raia Hadsell.
\newblock Progressive neural networks.
\newblock June 2016.
\newblock URL \url{http://arxiv.org/abs/1606.04671}.

\bibitem[Finn et~al.(2019)Finn, Rajeswaran, Kakade, and Levine]{Finn2019-ke}
Chelsea Finn, Aravind Rajeswaran, Sham Kakade, and Sergey Levine.
\newblock Online {Meta-Learning}.
\newblock In Kamalika Chaudhuri and Ruslan Salakhutdinov, editors,
  \emph{Proceedings of the 36th International Conference on Machine Learning},
  volume~97 of \emph{Proceedings of Machine Learning Research}, pages
  1920--1930. PMLR, 2019.
\newblock URL \url{http://proceedings.mlr.press/v97/finn19a.html}.

\bibitem[Gupta et~al.(2020)Gupta, Yadav, and Paull]{Gupta2020-of}
Gunshi Gupta, Karmesh Yadav, and Liam Paull.
\newblock {La-MAML}: Look-ahead meta learning for continual learning.
\newblock July 2020.
\newblock URL \url{http://arxiv.org/abs/2007.13904}.

\bibitem[Hendrycks and Gimpel(2016)]{DBLP:journals/corr/HendrycksG16c}
Dan Hendrycks and Kevin Gimpel.
\newblock A baseline for detecting misclassified and out-of-distribution
  examples in neural networks.
\newblock \emph{CoRR}, abs/1610.02136, 2016.
\newblock URL \url{http://arxiv.org/abs/1610.02136}.

\bibitem[Lee et~al.(2018)Lee, Lee, Lee, and Shin]{lee2018simple}
Kimin Lee, Kibok Lee, Honglak Lee, and Jinwoo Shin.
\newblock A simple unified framework for detecting out-of-distribution samples
  and adversarial attacks, 2018.

\bibitem[Hendrycks et~al.(2018)Hendrycks, Mazeika, and
  Dietterich]{DBLP:journals/corr/abs-1812-04606}
Dan Hendrycks, Mantas Mazeika, and Thomas~G. Dietterich.
\newblock Deep anomaly detection with outlier exposure.
\newblock \emph{CoRR}, abs/1812.04606, 2018.
\newblock URL \url{http://arxiv.org/abs/1812.04606}.

\bibitem[Roy et~al.(2021)Roy, Ren, Azizi, Loh, Natarajan, Mustafa, Pawlowski,
  Freyberg, Liu, Beaver, Vo, Bui, Winter, MacWilliams, Corrado, Telang, Liu,
  Cemgil, Karthikesalingam, Lakshminarayanan, and
  Winkens]{DBLP:journals/corr/abs-2104-03829}
Abhijit~Guha Roy, Jie Ren, Shekoofeh Azizi, Aaron Loh, Vivek Natarajan, Basil
  Mustafa, Nick Pawlowski, Jan Freyberg, Yuan Liu, Zachary Beaver, Nam Vo,
  Peggy Bui, Samantha Winter, Patricia MacWilliams, Gregory~S. Corrado, Umesh
  Telang, Yun Liu, A.~Taylan Cemgil, Alan Karthikesalingam, Balaji
  Lakshminarayanan, and Jim Winkens.
\newblock Does your dermatology classifier know what it doesn't know? detecting
  the long-tail of unseen conditions.
\newblock \emph{CoRR}, abs/2104.03829, 2021.
\newblock URL \url{https://arxiv.org/abs/2104.03829}.

\end{thebibliography}

\appendix

\section{Algorithms}

\subsubsection{Derivation of ELBO}
\label{app:elbo}

We assume a generative model with the following structure: 
the latent task process $Z_t$ is Markov 
\begin{equation}
P(Z_t|X_{<t}, Y_{<t}, Z_{t^{'} < t}) = P(Z_t|Z_{t-1})
\label{Eqn:PGM1}
\end{equation}
and the target variable $Y_t$ is conditionally independent of past observations $X_{<t}$ and $Y_{<t}$ given the current input observation $X_t$ and latent task $Z_t$ i.e. 
\begin{equation}
P(Y_t| X_{\leq t}, Y_{<t}, Z_{\leq t}) = P(Y_t|X_t, Z_t)
\label{Eqn:prediction}
\end{equation}
\textbf{Derivation of Variational Lower Bound:} Let's assume we are given a stream of observations $\tau := (x_{0:T}, y_{0:T})$. \footnote{\textbf{Notation}: We will use capital letters (e.g. $Z_t$ to denote random variables and small letters $z_t$ to denote realizations/observations of the random variable} We would like to approximate the (in general) intractable posterior distribution $P(Z_{\leq t}|X_{\leq t}, Y_{\leq t})$ using a distribution $q(Z_{\leq t}|X_{\leq t}, Y_{\leq t})$ from a class of tractable distributions $\mathcal{Q}$. We can find the best approximating $q$ by minimizing the KL divergence between the approximating distribution and the true posterior distribution 
\[
\min_{q \in Q} D_{KL}(q||P)
\]
where 
\begin{eqnarray}
 D_{KL}(q||P)\\
 & =  & E_{Z_{\leq t} \sim q} \left[ \log \frac{q(Z_{\leq t}|X_{\leq t}, Y_{\leq t})}{P(Z_{\leq t}|X_{\leq t}, Y_{\leq t})} \right] \nonumber \\
 & = & E_{Z_{\leq t} \sim q} \left[ \log \frac{q(Z_{\leq t}|X_{\leq t}, Y_{\leq t})}{P(Z_{\leq t}, X_{\leq t}, Y_{\leq t})} \cdot P(X_{\leq t}, Y_{\leq t}) \right] \nonumber \\
& = & E_{Z_{\leq t} \sim q} \left[ \log \frac{q(Z_{\leq t}|X_{\leq t}, Y_{\leq t})}{P(Z_{\leq t}, X_{\leq t}, Y_{\leq t})} \right] + \log P(X_{\leq t}, Y_{\leq t}) 
\end{eqnarray}

Rearranging terms, 
\begin{eqnarray}
\log P(X_{\leq t}, Y_{\leq t}) & = & D_{KL}(q||P) + E_{Z_{\leq t} \sim q} \left[ \log \frac{P(Z_{\leq t}, X_{\leq t}, Y_{\leq t})}{q(Z_{\leq t}|X_{\leq t}, Y_{\leq t})} \right] \\
& \geq & E_{Z_{\leq t} \sim q} \left[ \log \frac{P(Z_{\leq t}, X_{\leq t}, Y_{\leq t})}{q(Z_{\leq t}|X_{\leq t}, Y_{\leq t})} \right]
\end{eqnarray}
since $D_{KL}(q||P) \geq 0 $. That is, the log-likelihood of the observed trajectory $\tau$ is lower bounded by the evidence lower bound (ELBO) or variational lower bound  $\mathcal{L}(\tau; P, q)$ defined as follows
\begin{equation}
    \log P(\tau) \geq \underbrace{E_{Z_{\leq T} \sim q} \left[ \log \frac{P(Z_{\leq T}, x_{0:T}, y_{0:T})}{q(Z_{\leq T}|x_{0:T}, y_{0:T})} \right]}_{\mathcal{L}(\tau; P, q)} \label{Eq:ELBOdef}
\end{equation}

We further assume a factored form of $q$ (as in \cite{Krishnan2015-db})
\[
q(Z_0) \prod _{t=1}^{T} q( Z_t | Z_{t-1}, X_t, Y_t)
\]
Plugging into Eq (\ref{Eq:ELBOdef}), we get 
\begin{equation}
  \mathcal{L}(\tau; P, q) =   E_{Z_{\leq T} \sim q} \left[ \log \frac{P(Z_{\leq T},x_{\leq T}, y_{\leq T})}{q(Z_0) \prod_{\tp=1}^{T} q( Z_{\tp} | Z_{\tp-1}, x_{\tp}, y_{\tp}))} \right]
\end{equation}
Using the generative model assumptions, the numerator can be factorized as 
\begin{eqnarray}
P(Z_{\leq T}, X_{\leq T}, Y_{\leq T}) & = & \prod_{t} P(Z_t, X_t, Y_t| Z_{<t}, X_{<t}, Y_{<t}) \\
& = & \prod_{\tp} P(Y_\tp| Z_\tp, X_\tp) P( X_\tp) P(Z_\tp| Z_{\tp-1}) 
\end{eqnarray}
Plugging in, the ELBO $\mathcal{L}(\tau; P, q)$ is given by 
\[
E_{Z_{\leq T} \sim q} \left[ \log \frac{P(Z_0)P(y_0|x_0, Z_0)P(x_0) \prod_{t=1}^{T} P(y_\tp| Z_t, x_t) P( x_t) P(Z_t| Z_{t-1})}{q(Z_0) \prod_{t=1}^{t} q( Z_{t} | Z_{t-1}, x_{t}, y_{t})} \right]
\]

The $P(x_t)$ terms can be treated as a constant in the ELBO which then simplifies to 

\begin{equation}
\begin{split}
 \mathcal{L}(\tau; P, q) & = 
    E_{Z_t \sim q} \left[ \sum_{t=0}^{T} \log P(y_t| x_t, Z_t)\right] \\
    & ~~~~~~~- \sum_{t=1}^{T} E_{Z_{t-1} \sim q} \left[ D_{KL}(q( Z_{t} | Z_{t-1}, x_{t}, y_{t}) || P(Z_t|Z_{t-1})\right] \\
    &~~~~~~~ - D_{KL}(q(Z_0) || P(Z_0))
\end{split}
    \label{Eq:ELBOfinal}
\end{equation}
The form of the ELBO is similar to that in the original VAE in that the first term measures the  average reconstruction error using the model $P$ and the latent variables sampled from $q$, while the second and third terms act as a regularization constraining KL-divergence between the approximating posterior distribution $q$ and the prior.

\subsection{Implementation details}
For MAML prior, we train an ensemble of size 4. Similar to \cite{lakshminarayanan2017simple}, we use all available task data but with random mini-batch during the training. For each basis model, it will adapt to one specific task from MAML prior. Each basis model will also have an ensemble of size 4. Each NN in the ensemble will adapt from one of the MAML prior ensemble. We use one layer LSTM followed by a two layers projection head to predict the next $Y$. Each layer has 128 nodes. We use batch size of 32 to train both MAML and basis models. For Mob training, $\theta$ and $\phi$ are both parameterized by three layers fully-connected network with 128 hidden units. They output the $Z_t$ in the 32-dimensional latent space $Z$. To create new task, we use OOD buffer size T=20 and temperature $\eta=10$. To optimize our model, we use Adam with learning rate 1e-4. During the online training, we alternatively update between the (i) (basis models) and (ii),(iii) (mixture network $w_\theta(Z)$,the prior task model $p_\theta(Z_t|Z_{t-1})$) to stabilize the learning. We trained on a 4-GPU machine.

\begin{figure}[h!]
\begin{center}
\includegraphics[scale=0.3]{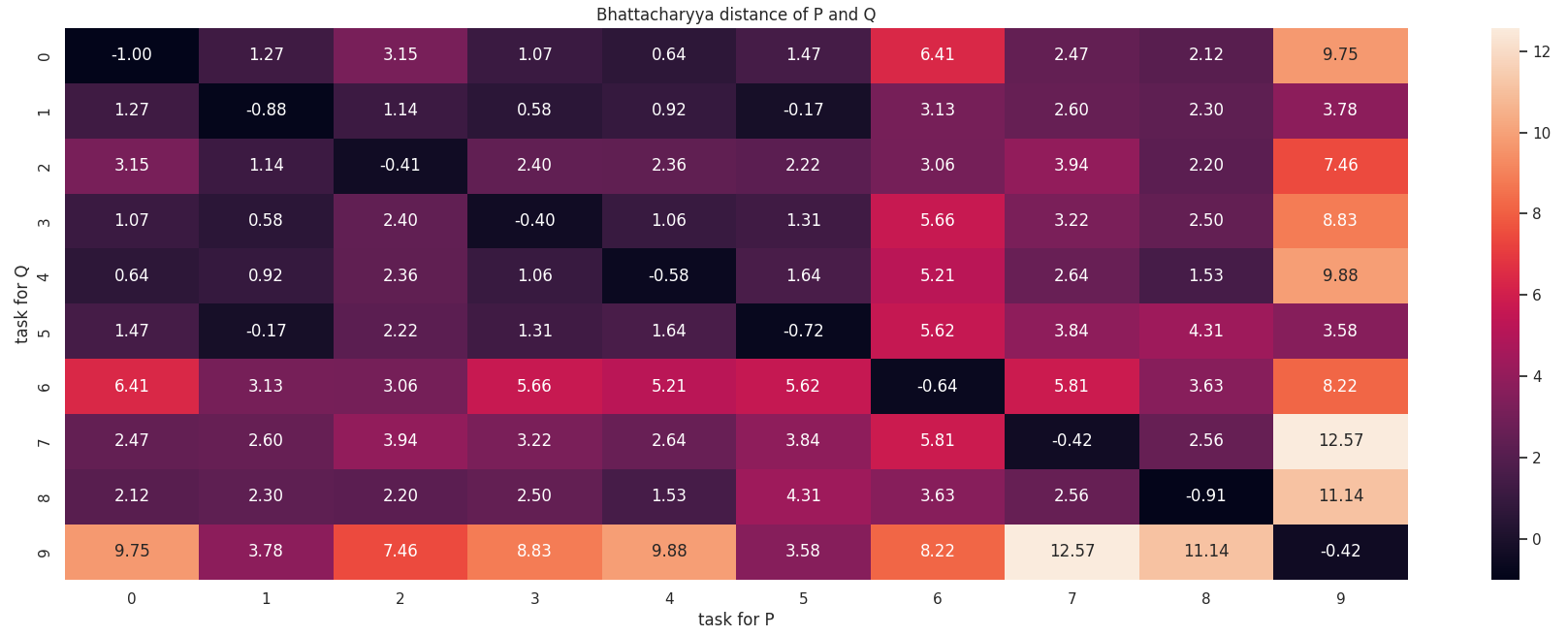}
\end{center}
\vspace{.05in}
\caption{\textbf{Regression domain:} Bhattacharyya distance between tasks.}
\label{fig:bhatta_dist}
\end{figure}

\begin{table}[h]
\caption{Task partition for experiments: the tasks listed under segmented are present both in the segmented and unsegmented offline datasets.}
\vspace{0.05in}
\centering
\begin{tabular}{l C{1.5cm} *{2}{C{3.5cm}}}
 \toprule
\textbf{Domain} & \textbf{Partition} & \textbf{Segmented} &\textbf{Unsegmented} \\
 \midrule
& 1 & 0, 5 & 1, 2, 6 \\ \cmidrule(l){2-4}
Regression & 2 & 4, 7 &  5, 6, 9 \\ \cmidrule(l){2-4}
& 3 & 2, 7 & 0, 3, 9 \\
\midrule
\midrule
& 1 & bt, ff & btff, bfft \\ \cmidrule(l){2-4}
HalfCheetah Foot & 2 & bfff, btft & btbf, ftff \\ \cmidrule(l){2-4}
& 3 & btbf, ftff & bfff, btft \\
\midrule
\midrule
& \multirow{2}{*}{1} & \multirow{2}{*}{medium slope, flat ice} & medium ice slope, mixed slope\\ \cmidrule(l){2-4}
\multirow{2}{*}{HalfCheetah Ice-Slope} & \multirow{2}{*}{2} & \multirow{2}{*}{mixed slopes, flat ice} & medium ice slope, medium slope \\ \cmidrule(l){2-4}
& \multirow{2}{*}{3} & medium ice slope, medium slope & \multirow{2}{*}{mixed slopes, flat ice}\\
\bottomrule
\end{tabular}
\label{tab:all_task_seg}
\end{table}

\section{Additional Experimental Results}

\begin{figure}[h!]
\begin{center}
\includegraphics[scale=0.35]{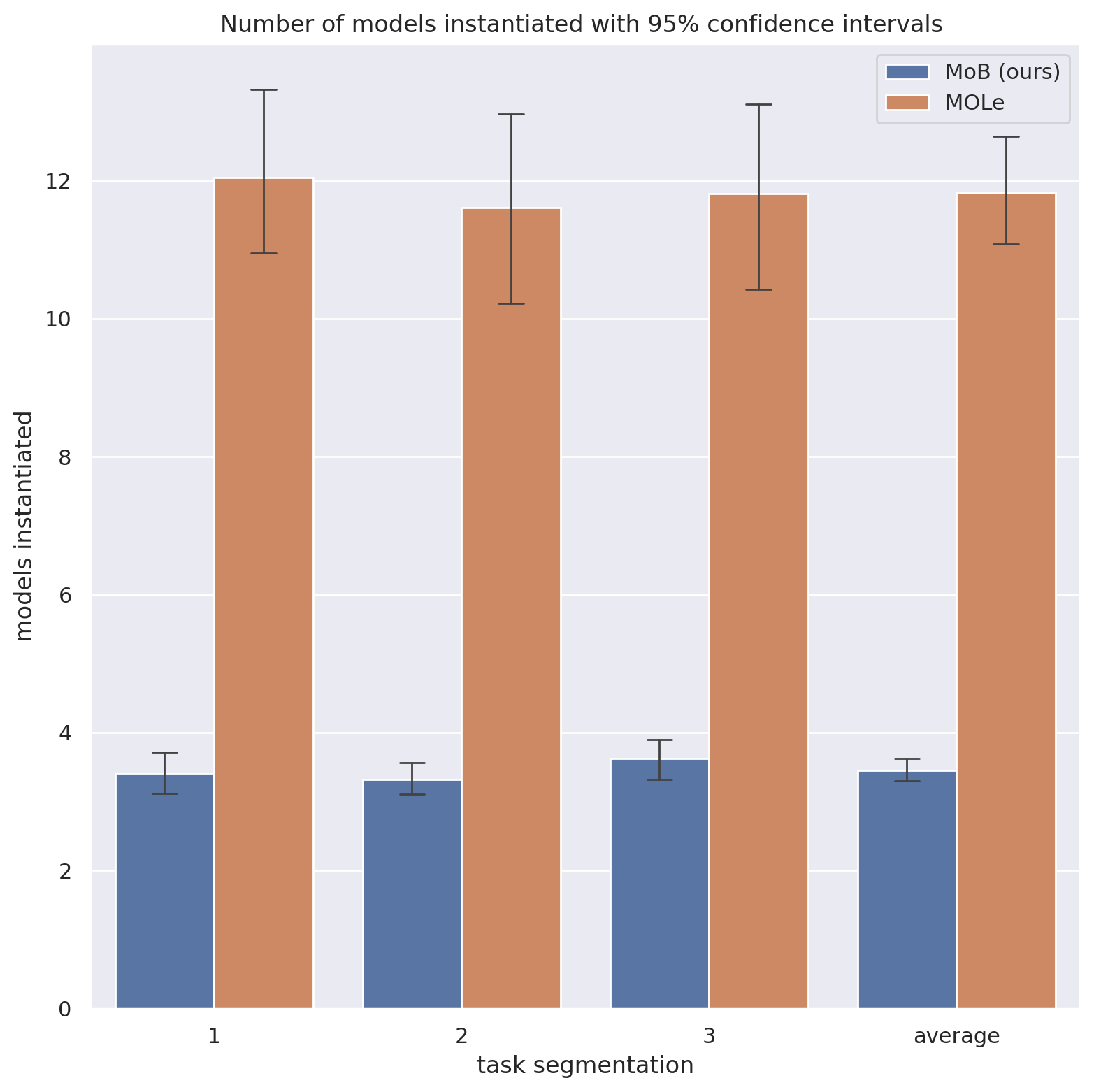}
\end{center}
\caption{Number of models instantiated by MoB and MOLe, the two algorithms we evaluate that create additional models to handle distribution shifts, with 95\% confidence intervals. For each task partition MoB and MOLe were each evaluated on the same 500 trajectories (10 seeds * 50 trajectories/seed) mentioned in Figure \ref{fig:reg_losses}. For each algorithm for each task partition, the means and 95\% confidence intervals of the mean losses for each seed were computed and are represented by the bar plots and interval lines, respectively. 'average' depicts the arithmetic mean of the results over the three task partitions.}
\label{fig:reg_nbasis}
\end{figure}

\begin{figure}[h!]
\begin{center}
\includegraphics[scale=0.35]{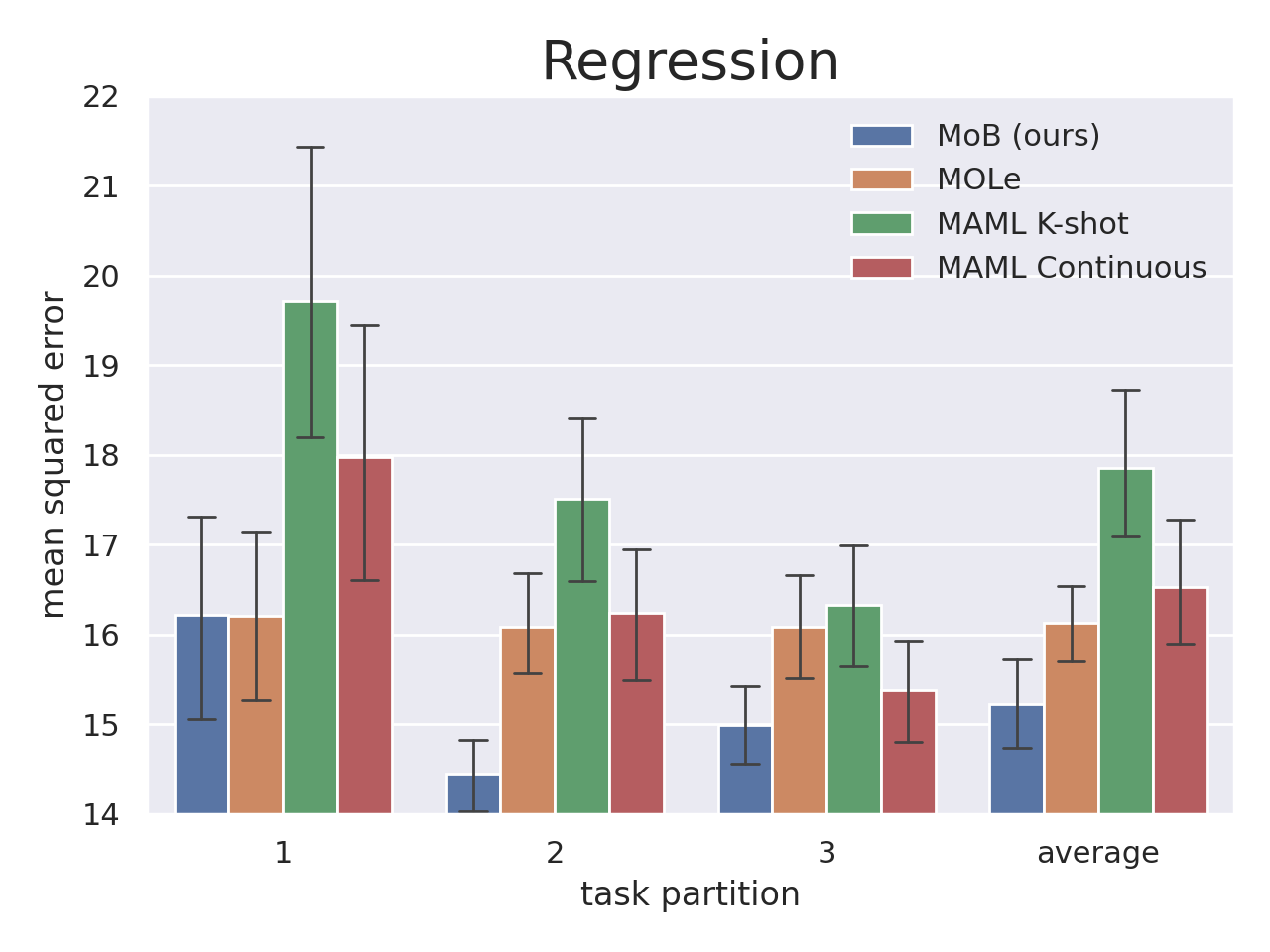}
\end{center}
\caption{Mean squared errors and mean absolute deviations averaged over 10 seeds for each task partition. For each task partition 50 new trajectories were randomly sampled from an unseen test set. Each algorithm was evaluated on these trajectories to obtain a mean squared error and mean absolute deviation for this particular seed and partition. Lastly for each algorithm for each task partition, the means and 95\% confidence intervals of the mean losses for each seed were computed and are represented by the bar plots and interval lines, respectively. 'average' depicts the arithmetic mean of the results over the three task partitions.}
\label{fig:reg_losses}
\end{figure}

\begin{figure}[h!]
\begin{center}
\includegraphics[scale=0.35]{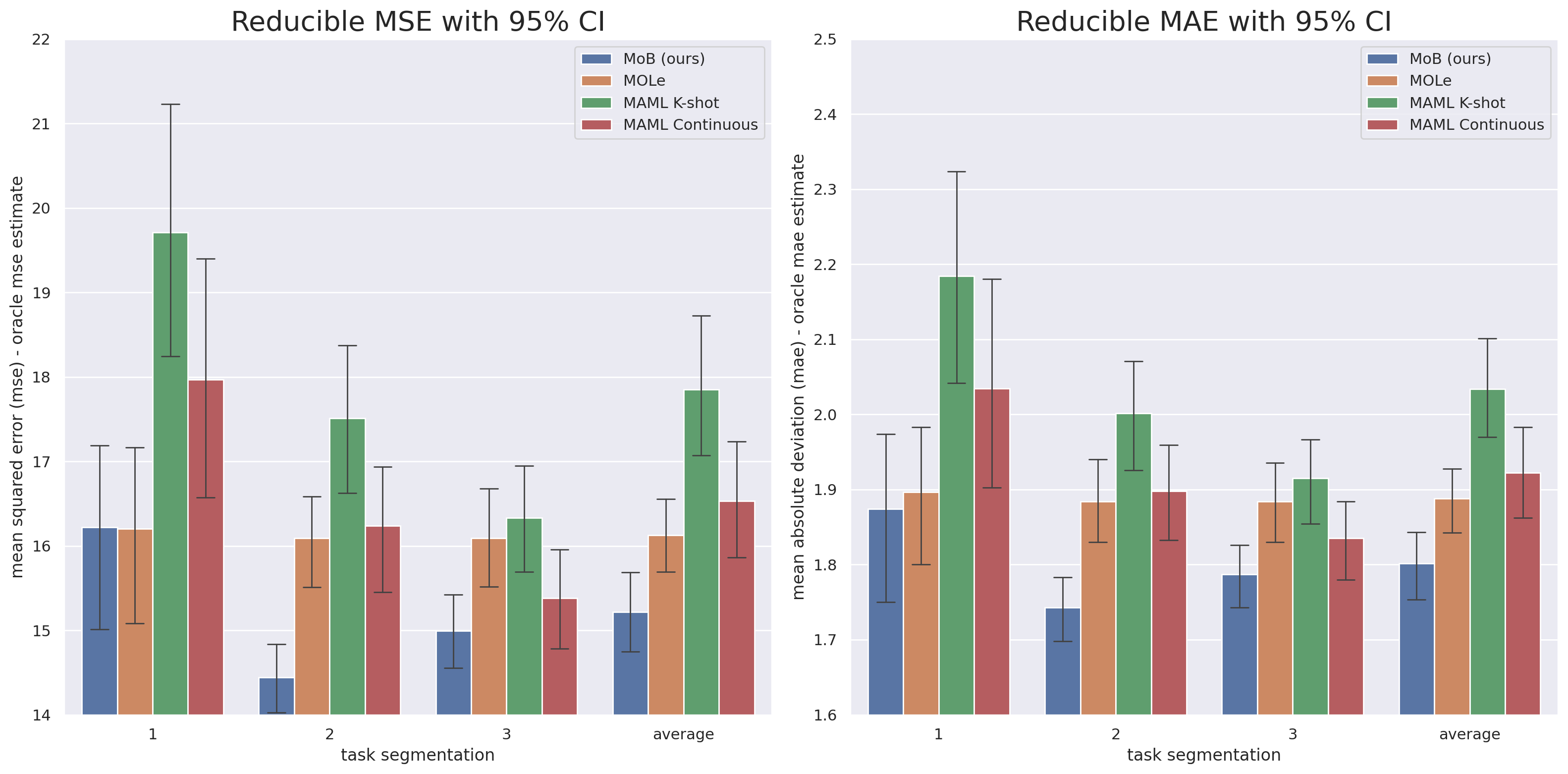}
\end{center}
\caption{Mean squared errors and mean absolute deviations reduced by Monte Carlo estimates of the irreducible MSEs and MAEs, respectively, averaged over 10 seeds for each task partition. In essence the values shown are Monte Carlo estimates of the reducible errors attained by each algorithm. The method used to compute these results is the same as explained in the caption of \ref{fig:reg_losses}.}
\label{fig:reg_mcerror}
\end{figure}

\begin{table}[h]
 \caption{Regression task losses averaged over task partitions (mean, standard deviation)}
 \label{table:reg_losses}
\vspace{0.05in}
\begin{center}
\begin{tabular}{l *{4}{C{2.0cm}}}

 \toprule
 \textbf{Algorithm} & \textbf{MSE} & \textbf{MAE} & \textbf{Reducible MSE} & \textbf{Reducible MAE}\\
 \midrule
MOLe            & 22.38, \ 1.22 & 3.73, \ 0.12 & 16.13, \ 1.19 & 1.89, \ 0.12\\
MAML K-shot     & 24.10, \ 2.32 & 3.88, \ 0.20 & 17.85, \ 2.32 & 2.03, \ 0.20\\
MAML Continuous & 22.78, \ 1.99 & 3.77, \ 0.17 & 16.53, \ 1.98 & 1.92, \ 0.17\\
 \midrule
MoB (ours)            & \textbf{21.47, \ 1.37} & \textbf{3.64, \ 0.13} & \textbf{15.22, \ 1.43} & \textbf{1.80, \ 0.14}\\
 \bottomrule
\end{tabular}
\end{center}
\end{table}

\begin{figure}[h!]
\begin{center}
\includegraphics[width=\textwidth]{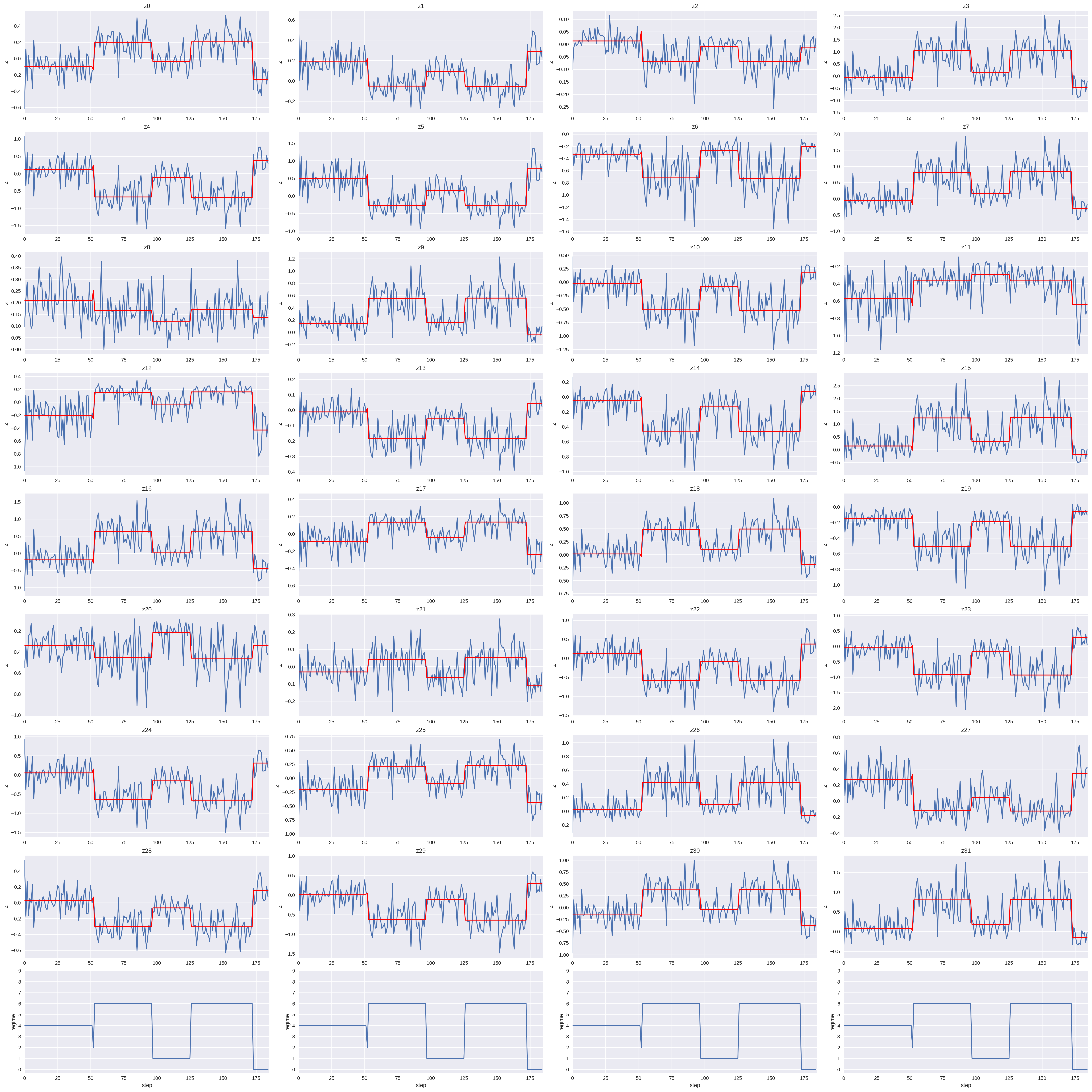}
\end{center}
\caption{Following the Offline training phase on the Regression domain, these are activations of all 32 $Z$ components as the active task varies according to the task plots at the bottom. The plots in each column are aligned so the activations at any time $t$ vertically line up across subplots with other activations and the active task at time $t$.}
\label{fig:9x4}
\end{figure}

\begin{figure}[h!]
\begin{center}
\includegraphics[width=\textwidth]{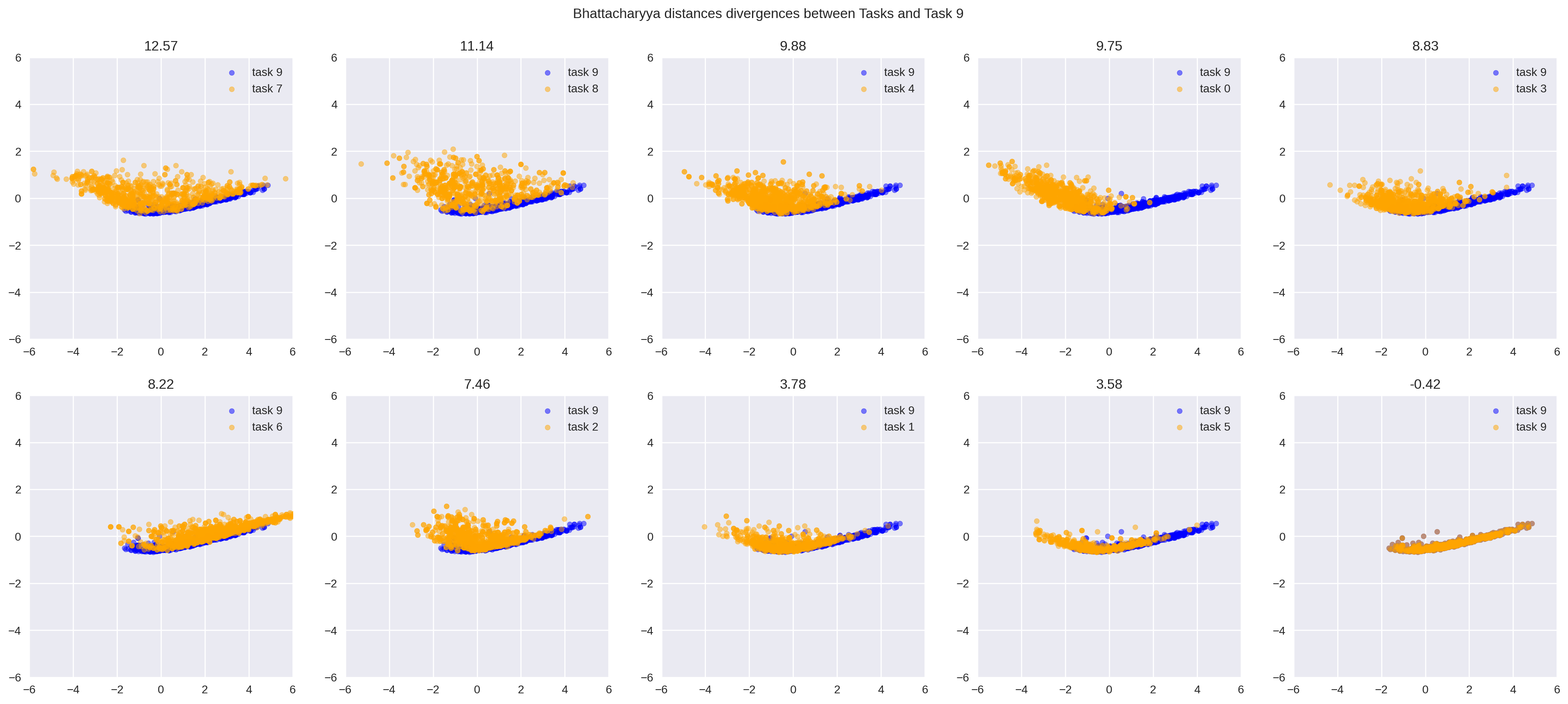}
\end{center}
\caption{PCA embedding of latent space $Z$ of regression dataset. Bhattacharyya distance between tasks is above each subplot.}
\label{fig:pca_bhatta_all}
\end{figure}

\begin{figure}[h!]
    \begin{subfigure}{0.45\textwidth}
        \centering
        \includegraphics[scale=0.125]{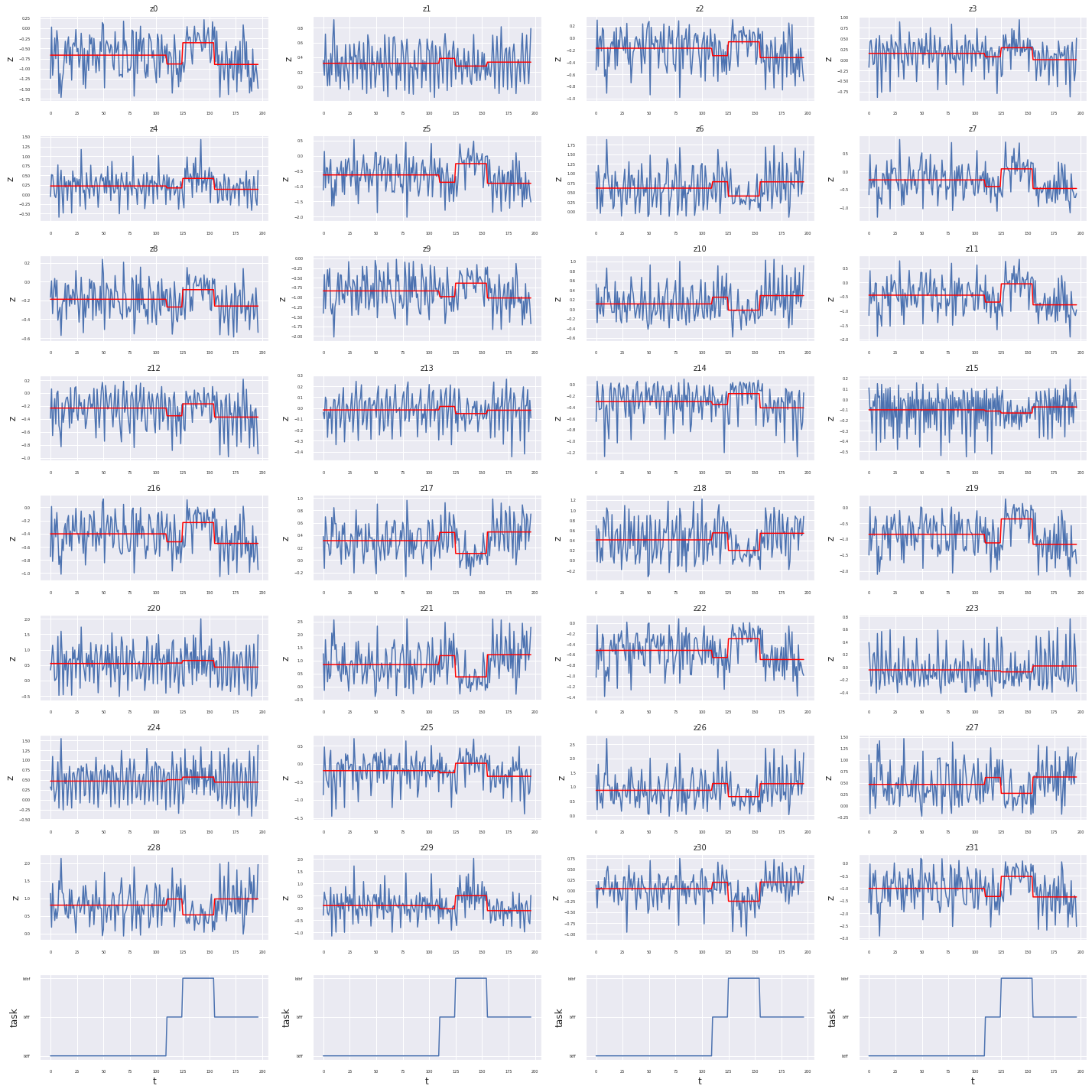}
        \caption{Following the Offline training phase on the HalfCheetah foot, these are activations of all 32 $Z$ components as the active task varies according to the task plots at the bottom.}
        \label{fig:foot_all_z}
    \end{subfigure}
    \hfill
    \begin{subfigure}{0.45\textwidth}
        \centering

        \includegraphics[scale=0.15]{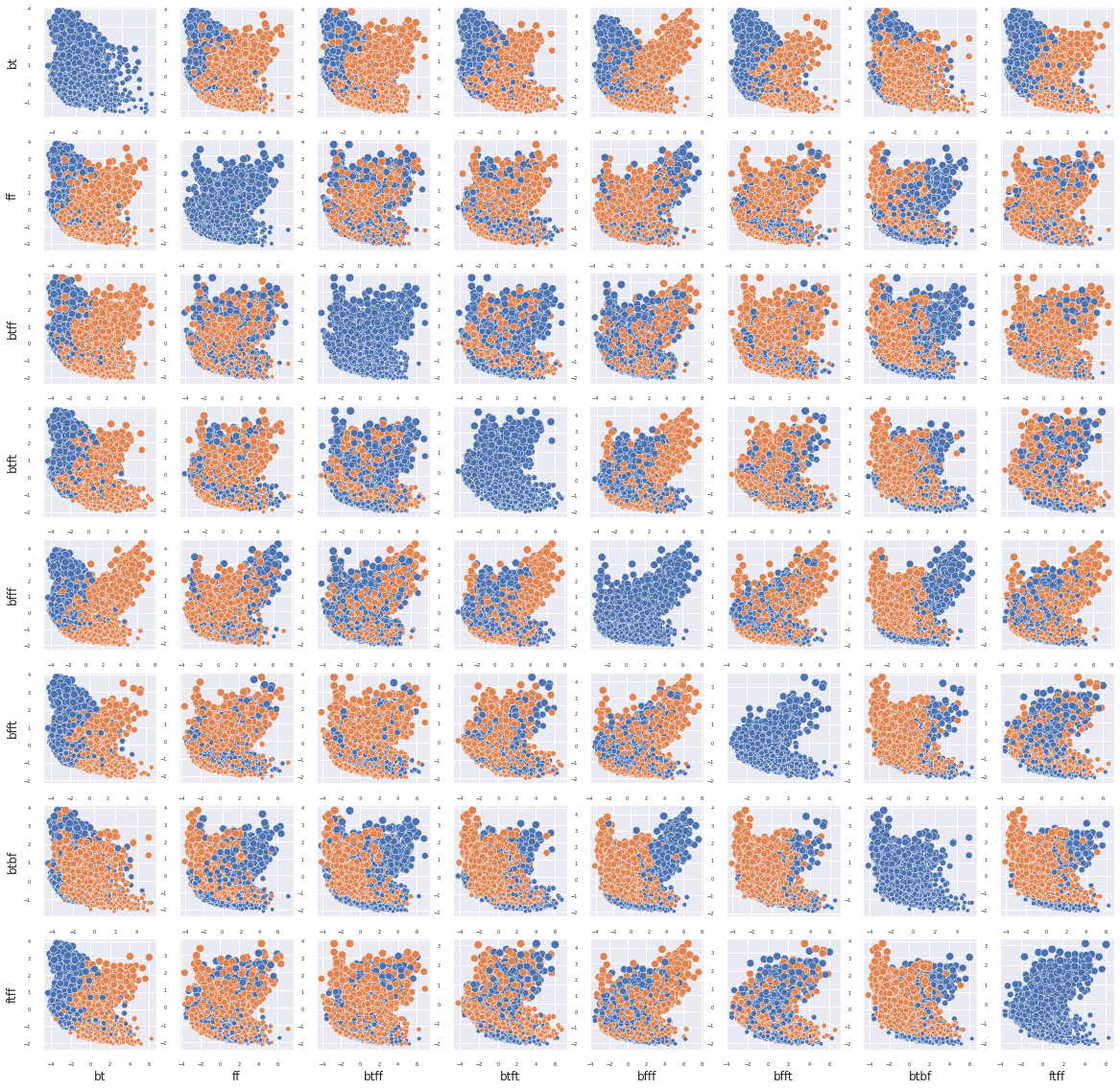}
        \caption{8x8 task pairwise plot over PCA embedding of latent space $Z$ in HalfCheetah foot. First, task \textit{bt} and \textit{ff} distribution are quite different in the PCA embedding. Though they overlap on some region, it might because they still have some commonality between tasks. Tasks containing clipping \textit{bt} is closer to task \textit{bt}. Similarly, any task containing clipping \textit{ff} or \textit{ft} is closer to \textit{ff}.}. 
        \label{fig:foot_pca}
    \end{subfigure}
    \caption{Latent space $Z$ of HalfCheetah foot}
\end{figure}

\begin{figure}[h!]
    \begin{subfigure}{0.45\textwidth}
        \centering
        \includegraphics[scale=0.125]{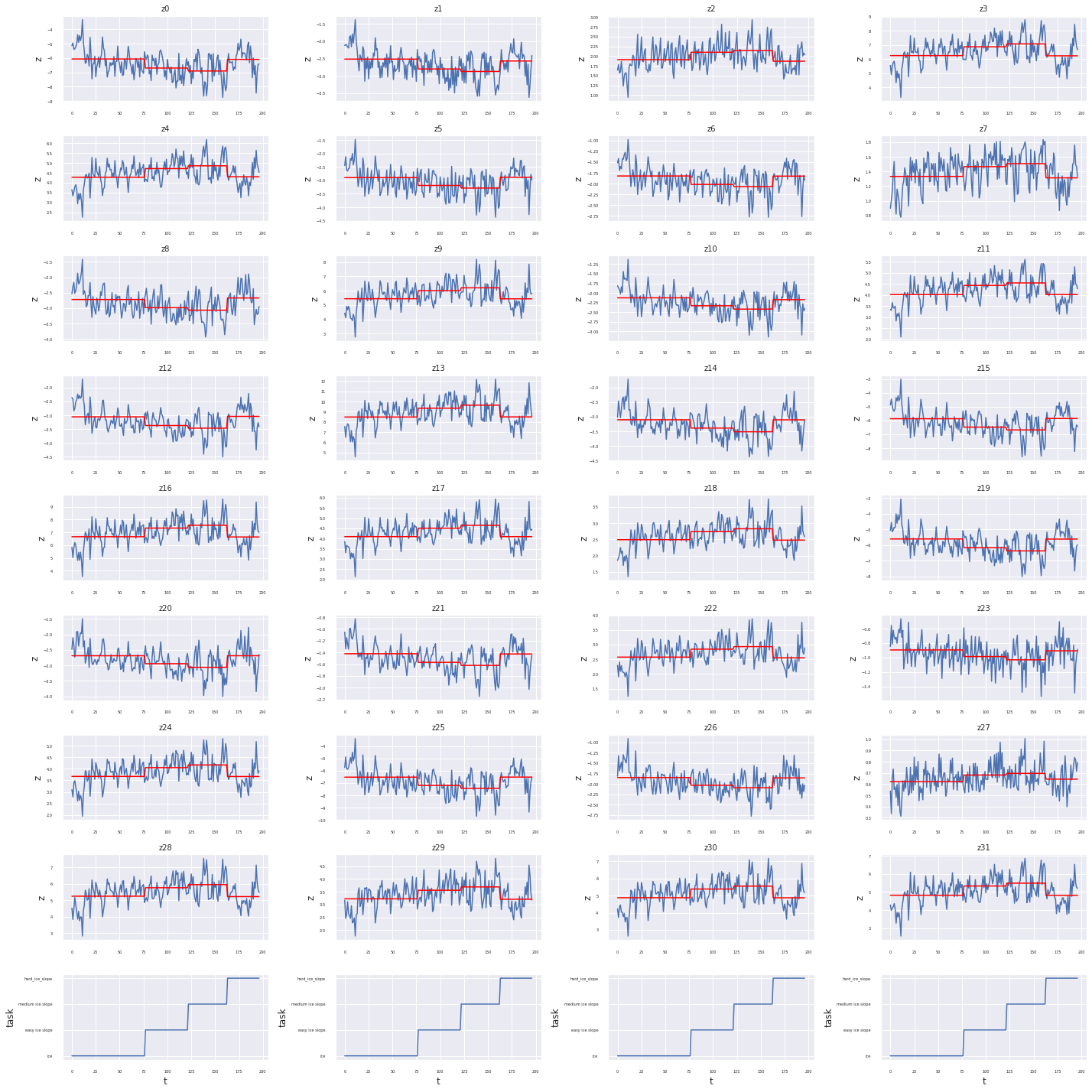}
        \caption{Following the Offline training phase on the HalfCheetah ice-slope, these are activations of all 32 $Z$ components as the active task varies according to the task plots at the bottom.}
        \label{fig:hill_all_z}
    \end{subfigure}
    \hfill
    \begin{subfigure}{0.45\textwidth}
        \centering

        \includegraphics[scale=0.15]{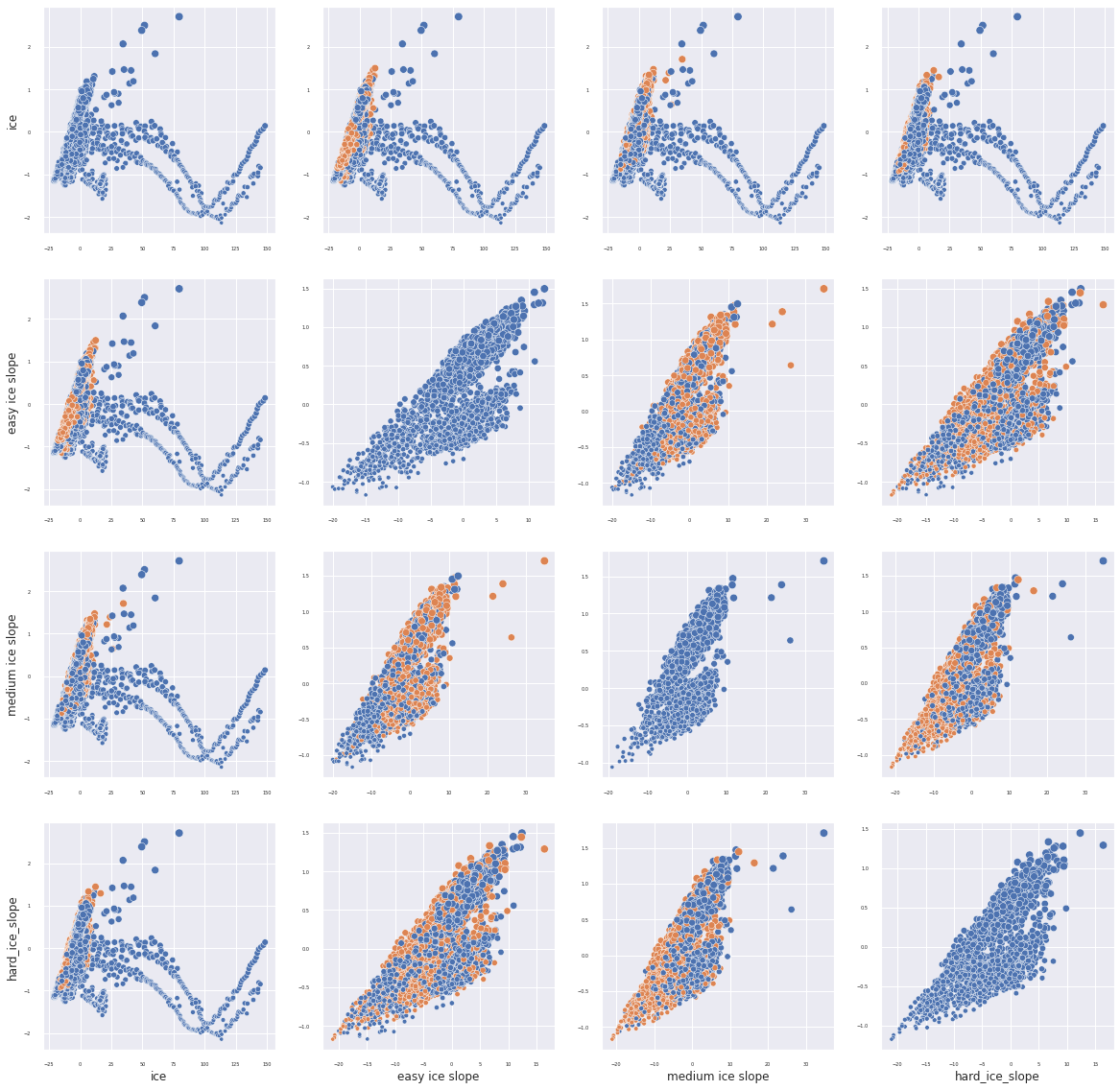}
        \caption{8x8 task pairwise plot over PCA embedding of latent space $Z$ in HalfCheetah ice-slope. One interesting observation is that medium ice-slope and hard ice-slope distribution are different subsets of easy ice-slope distribution. We conjugate that tasks are similar and different level of ice-slopes can be expressed by different subset of easy ice-slope latent space.}
        \label{fig:hill_pca}
    \end{subfigure}
    \caption{Latent space $Z$ of HalfCheetah ice-slope}
\end{figure}

\end{document}